\documentclass{article} 
\usepackage{iclr2024_conference,times}

\usepackage{amsmath,amsfonts,bm}

\def\eqref#1{equation~\ref{#1}}

\def\1{\bm{1}}

\DeclareMathAlphabet{\mathsfit}{\encodingdefault}{\sfdefault}{m}{sl}
\SetMathAlphabet{\mathsfit}{bold}{\encodingdefault}{\sfdefault}{bx}{n}

\newcommand{\bx}{\boldsymbol{x}}

\newcommand{\cfg}{noise-damped classifier-free guidance}
\newcommand{\inflateconv}{dispersed convolution\ }

\newcommand{\CUT}[1]{}

\usepackage{amssymb}
\usepackage{bbding}
\usepackage{pifont}
\usepackage{hyperref}
\usepackage{url}
\usepackage{graphicx}
\usepackage{booktabs}
\usepackage{xcolor}
\usepackage{multirow}
\usepackage{wrapfig}
\usepackage{subfig}
\usepackage[super]{nth}
\usepackage[capitalize]{cleveref}
\crefname{section}{Sec.}{Secs.}
\Crefname{section}{Section}{Sections}
\Crefname{table}{Table}{Tables}
\crefname{table}{Tab.}{Tabs.}

\title{ScaleCrafter: Tuning-free High{\color{cyan}er}-Resolu-\\tion Visual Generation with Diffusion Models}
\iclrfinalcopy

\author{Yingqing He\thanks{Equal Contribution}$^{\;\,1,3}$, 
Shaoshu Yang$^{*2,3}$, 
Haoxin Chen$^3$, 
Xiaodong Cun$^3$, 
Menghan Xia$^3$,\\
\textbf{Yong Zhang}\thanks{Corresponding Authors}$\;\,^3$, 
\textbf{Xintao Wang}$^3$, 
\textbf{Ran He}$^2$, 
\textbf{Qifeng Chen}$^{\dagger1}$, 
\textbf{Ying Shan}$^3$\\
\\ 
\small $^{1}$Hong Kong University of Science and Technology\\
\small $^{2}$Chinese Academy of Sciences \\
\small $^{3}$Tencent AI Lab
}

\begin{document}

\maketitle

\begin{abstract}
In this work, we investigate the capability of generating images from pre-trained diffusion models at much higher resolutions than the training image sizes. 
In addition, the generated images should have arbitrary image aspect ratios.
When generating images directly at a higher resolution, $1024 \times 1024$, with the pre-trained Stable Diffusion using training images of resolution $512 \times 512$, we observe persistent problems of object repetition and unreasonable object structures. 
Existing works for higher-resolution generation, such as attention-based and joint-diffusion approaches, cannot well address these issues.
As a new perspective, we examine the structural components of the U-Net in diffusion models and identify the crucial cause as the limited perception field of convolutional kernels. 
Based on this key observation, we propose a simple yet effective \textit{re-dilation} that can dynamically adjust the convolutional perception field during inference.
We further propose the \inflateconv and \cfg, which can enable \textit{ultra-high-resolution} image generation (\textit{e.g.,} $4096 \times 4096$).
Notably, our approach \textit{does not require any training or optimization}. 
Extensive experiments demonstrate that our approach can address the repetition issue well and achieve state-of-the-art performance on higher-resolution image synthesis, especially in texture details. 
Our work also suggests that a pre-trained diffusion model trained on low-resolution images can be directly used for high-resolution visual generation without further tuning, which may provide insights for future research on ultra-high-resolution image and video synthesis.
More results are available at the project website: \href{https://yingqinghe.github.io/scalecrafter/}{https://yingqinghe.github.io/scalecrafter/}.
\end{abstract}

\section{Introduction}

\begin{figure}[h]
  \centering
  \includegraphics[width=1.0\textwidth]{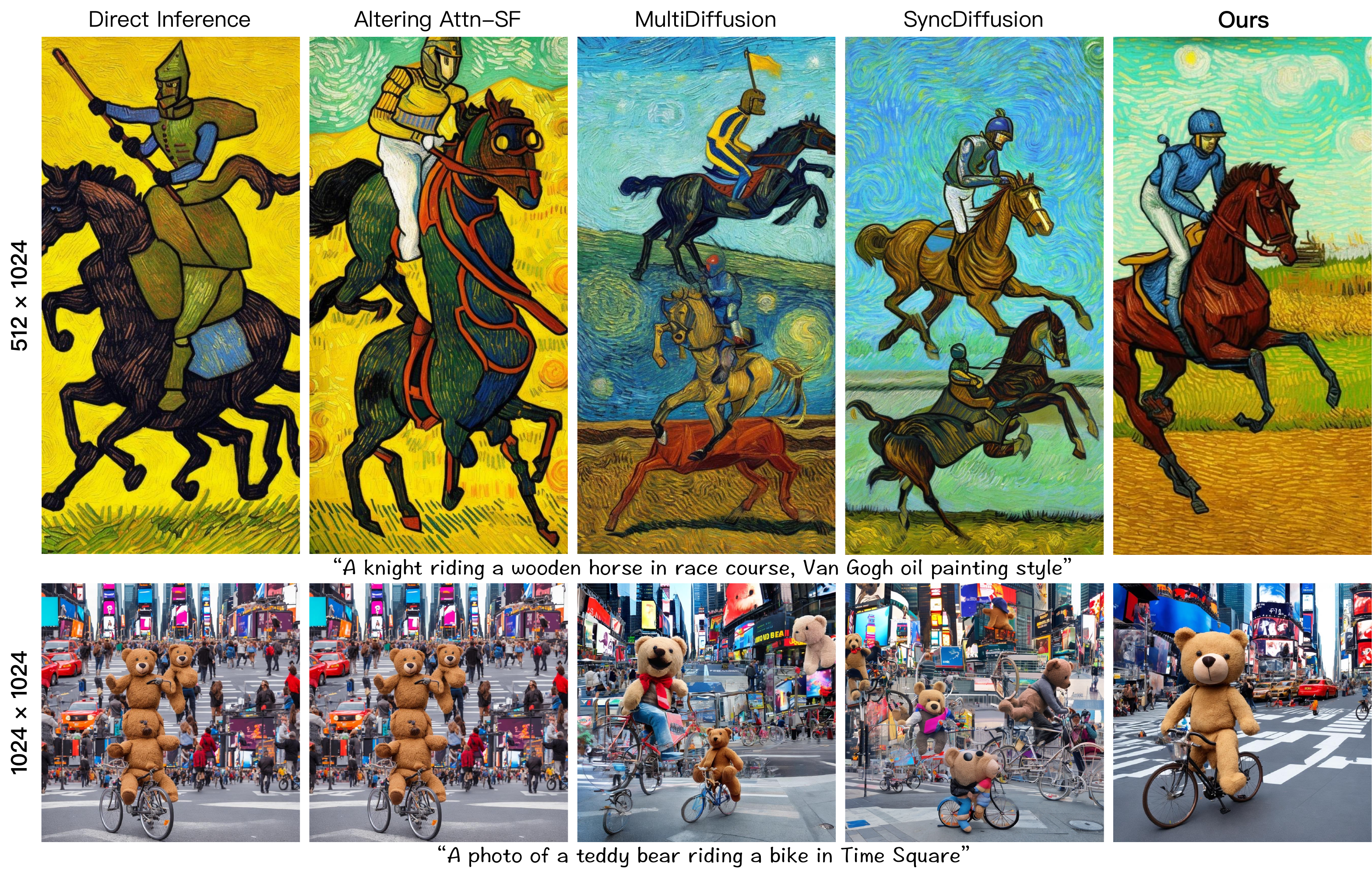}
  \caption{
  Structure repetition issue of higher-resolution generation (Train:  512$^2$; Inference: 512$\times$1024 and 1024$^2$).
  %
  Altering the scaling factor of attention~\citep{trainfree-variablesize}, and joint diffusion approaches including MultiDiffusion~\citep{multidiffusion} and SyncDiffusion~\citep{syncdiffusion} fails to address this problem.
  While our simple \textit{re-dilation} successfully solves this problem and yields structure and semantic correct images, and at meanwhile \textit{require no optimization and tuning cost}.
  }
  \label{fig:problem}
\end{figure}

\begin{figure}[t]
  \centering
  \includegraphics[width=0.9\textwidth]{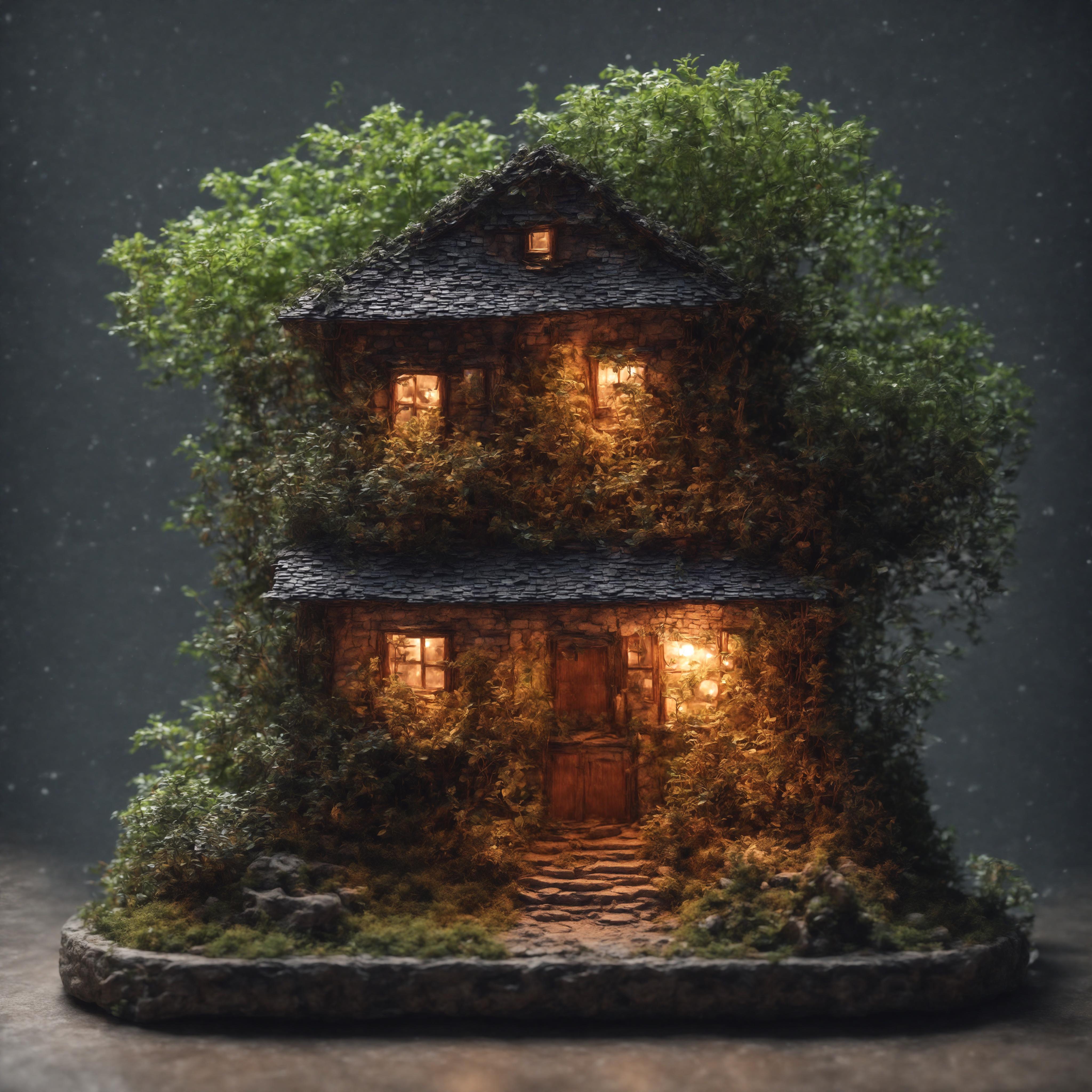}
  \caption{
  Our method can generate $4096 \times 4096$ images, 16$\times$ higher than the training resolution.
  }
  \label{fig:teaser}
\end{figure}

In recent two years, the rapid development of image synthesis has attracted tremendous attention from both academia and industry, especially the most popular text-to-image generation models, such as Stable Diffusion (SD)~\citep{ldm}, SD-XL~\citep{sdxl}, Midjourney~\citep{Midjourney}, and IF~\citep{IF}. 
However, the highest resolution of these models is $1024 \times 1024$, which is far from the demand of applications such as advertisements.  

Directly sampling an image with a resolution beyond the training image sizes of those models will encounter severe object repetition issues and unreasonable object structures. 
As shown in Fig.~\ref{fig:problem}, when using a Stable Diffusion (SD) model trained on images of $512 \times 512$, to sample images of $ 512 \times 1024$ and  $\1024 \times 1024$ resolutions, the object repetition appears. 
The larger the image size, the more severe the repetition. 

A few methods attempt to generate images with a larger size than the training image size of SD, \textit{e.g.,} Multi-Diffusion~\citep{multidiffusion} and SyncDiffusion~\citep{syncdiffusion}. 
In Multi-Diffusion, images generated from multiple windows are fused using the averaged features across the windows in all the reverse steps. 
While SyncDiffusion improves the style consistency of Multi-Diffusion by using an anchor window. 
However, they focus on the smoothness of the overlap region and cannot solve the repetition issue, as shown in Fig.~\ref{fig:problem}. 
Most recently, \citep{trainfree-variablesize} studies the SD adaptation for variable-sized image generation through the view of attention entropy. 
However, their method has a negligible effect on the object repetition issue when increasing the inference resolution.

%
%

To investigate the pattern repetition, we sample a set of images of $1024^2$ and $512^2$ from the pre-trained SD model trained with $512^2$ images for comparison. 
Zooming in on the images, we observe that the images of $1024^2$ have no blur effects and the image quality does not degenerate like the Bilinear upsampling, though their object structures become worse.  
It indicates that the pre-trained SD model has the potential to generate higher-resolution images without sacrificing image definition.  

We then delve into the structural components of SD to analyze their influence, \textit{e.g.,} convolution, self-attention, cross-attention, etc.  
Surprisingly, when we change convolution to dilated convolution in the whole U-Net using the pre-trained parameters, the overall structure becomes reasonable, \textit{i.e.,} the object repetition disappears. 
However, the repetition happens to local edges. 
We then carefully analyze \textit{where}, \textit{when}, and \textit{how} to apply the dilated convolution, \textit{i.e.,} the influence of U-Net blocks, timesteps, and dilation radius. 
Based on these studies, we propose a tuning-free dynamic re-dilation strategy to solve the repetition. 
However, as the resolution further increases (e.g., 16x), the decreased generated quality and denoising ability arise.
To tackle it, we then propose novel dispersed convolution and noise-damped classifier-free guidance for ultra-high-resolution generation. 

Our main contributions are as follows:
\begin{itemize}
    \item We observe that the primary cause of the object repetition issue is the limited convolutional receptive field rather than the attention token quantity, providing a new viewpoint compared to the prior works.
    \item Based on this observation, we propose the simple yet effective \textit{re-dilation} for dynamically increasing the receptive field during inference time. We also propose \textit{dispersed convolution} and \textit{\cfg} for ultra-high-resolution generation.
    \item We empirically evaluate our approach on various diffusion models, including different versions of Stable Diffusion, and a text-to-video model, with varying image resolutions and aspect ratios, demonstrating the effectiveness of our model.
\end{itemize}

\section{Related work}

\subsection{Text-to-Image Synthesis}
Text-to-image synthesis has gained remarkable attention recently due to its impressive generation performance \citep{imagen, dalle-2, muse, cogview2}.
Among various generative models, diffusion models are popular for their high-quality generation capabilities \citep{make-your-vid, lvdm, imagen-video}.
Following the groundbreaking work of DDPM \citep{ddpm}, numerous studies have focused on diffusion models for image generation \citep{glide, im-ddpm, adam, cascaded-dm, ldm, dalle-2, vqdm}.
In particular, Latent diffusion models (LDM) \citep{ldm} have become widely used, as their compact latent space improves model efficiency.
Subsequently, a series of Stable Diffusion (SD) models are open-sourced, building upon LDMs and offering high sample quality and creativity.
Despite their impressive synthesis capabilities, the resolution remains limited to the training resolution, \textit{e.g.}, 512$^2$ for SD 2.1 and 1024$^2$ for SD XL, necessitating a mechanism for higher resolution generation (\textit{e.g.}, 2K, 4K, etc.).

\subsection{High-resolution synthesis and adaptation}
High-resolution image synthesis is challenging due to the difficulty of learning from higher-dimensional data and the substantial requirement of computational resources. 
Prior work can mainly be divided into two categories: \textit{training from scratch} \citep{relay-diffusion,simple-diffusion, importance-diffusion} and \textit{fine-tuning} \citep{any-size-diffusion, difffit}.
Most recently, \citep{trainfree-variablesize} studies a training-free approach for variable-sized adaptation. However, it fails to tackle the case of higher-resolution generation.
Multi-Diffusion~\citep{multidiffusion} and SyncDiffusion~\citep{syncdiffusion} focus on smoothing the overlap region to avoid inconsistency between windows. 
However, object repetition still exists in their results.
MultiDiffusion can avoid repetition by using the conditions such as region and text from the user. 
However, those extra inputs are not available in the scenario of text-to-image generation. 
Differently, we propose a tuning-free method from the perspective of the network structure that can fundamentally solve the object repetition issue in higher-resolution image synthesis. 

\section{Method}
\begin{figure}[t]
  \centering
  \includegraphics[width=1\textwidth]{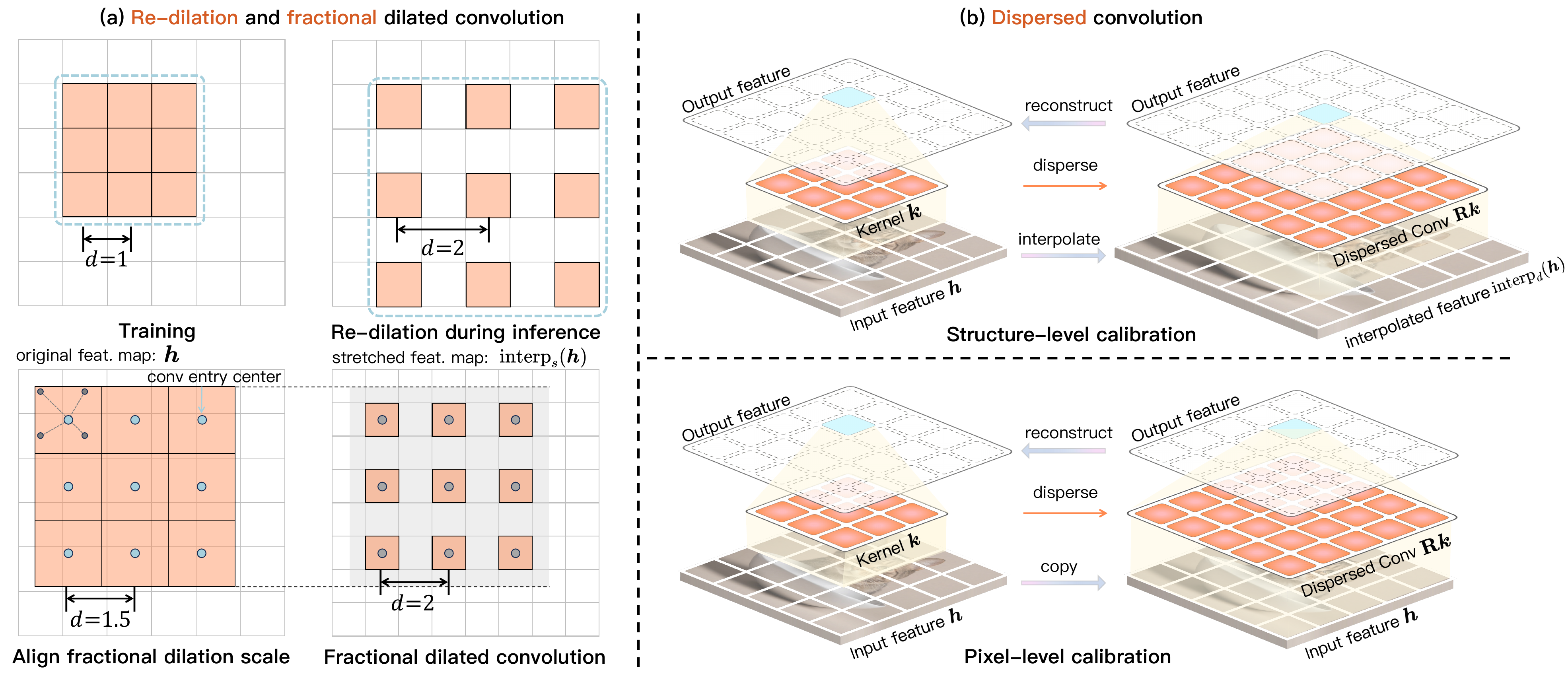}
  \caption{
  (a) The first row shows re-dilation. Given a pre-trained kernel trained on low-resolution data, we fix the parameters and insert spaces into kernel elements during test time. The second row shows fractional dilated convolution. For each entry of the convolution kernel, we compute the input feature with features near the kernel entry center with bilinear interpolation. This is equivalent to stretch input feature maps and uses a rounded-up dilation scale before the convolution operation.
  (b) Dispersed convolution can enlarge a pre-trained kernel with a specific scale. We use structure-level calibration to adapt to a new perception field when the input feature dimension is larger and use pixel-level calibration to preserve local information processing ability.
  }
  \label{fig:re-dilate}
\end{figure}

\subsection{Problem Formulation and Motivation}
Without the loss of generality, the formulation of this paper considers the $\epsilon$-prediction paradigm of diffusion models. Given a base diffusion model $\bm{\epsilon}_\theta(\cdot)$ parameterized by $\theta$. It is trained on a fixed pre-defined low-resolution image $\bx \in \mathbb{R}^{3 \times h \times w}$, our goal is to adapt the model to $\tilde{\bm{\epsilon}}_\theta(\cdot)$ in a training-free manner to synthesize higher resolution images $\tilde{\bx} \in \mathbb{R}^{3\times H\times W}$.

A previous work~\citep{trainfree-variablesize} attributes the degradation of performance to the change in the number of attention tokens and proposes to scale the features in the self-attention layer according to the input resolution. 
However, when applying it to generate $1024^2$ images, the object repetition is still there (see Fig.~\ref{fig:problem}). 
We observe that the local structure of each repetitive object seems reasonable and the unreasonable part is the object number when the resolution increases.
This encourages us to investigate whether the receptive field of any network component does not fit the larger resolution. 


Hence, we modify the components of the SD U-Net to analyze their influence, such as attention, convolution, normalization, etc. 
We develop \emph{re-dilation} to dissect the effect of the receptive field of attention and convolution, respectively.
Re-dilation aims to adjust the network receptive field on a higher-resolution image to maintain the same as the original lower-resolution generation.
For re-dilated attention, we partition the feature map into slices with each slice collected via a feature dilation having the same token quantity as training. 
Then, these slices are fed into the QKV attention in parallel, after which they are merged into the original arrangement. Details are illustrated in the supplementary.
However, maintaining the receptive field of attention yields results with indistinguishable differences compared with direct inference.
Differently, when increasing the receptive field of convolution in all blocks of the U-Net, fortunately, we observe that the number of objects is correct though there are many artifacts such as noisy background and repetitive edges.
Based on the observation, we then develop a more elaborate re-dilation strategy considering where, when, and how to apply dilated convolution. 

\subsection{Re-dilation}
Note that we ignore the feature channel dimension and convolution bias for simplicity in the following. Considering a hidden feature $\bm{h} \in \mathbb{R}^{m \times n}$ before a convolution layer $f_{\bm{k}}(\cdot)$ of the network. Given the convolution kernel $\bm{k} \in \mathbb{R}^{r\times r}$ and the dilation operation $\Phi_d(\cdot)$ with factor $d$. The dilated convolution $f_{\bm{k}}^d(\cdot)$ is computed with
\begin{equation}
f_{\bm{k}}^d(\bm{h}) = \bm{h} \circledast \Phi_d(\bm{k}), \; (\bm{h} \circledast \Phi_d(\bm{k}))(o) = \sum_{s + d \cdot t = p} \bm{h}(p) \cdot \bm{k}(q),
\label{eq:dilated}
\end{equation}
where $o,p,q$ are spatial locations used to index the feature and kernel, $\circledast$ denotes convolution operation. 
Notably, different from traditional dilated convolutions, which share a common dilation factor during training and inference. Our proposed approach dynamically adjusts the dilation factor \emph{only in inference time}, leading us to term it as \textit{re-dilation}.
Since the dilation factor can only be an integer, traditional dilated convolution cannot address a fractional multiple of the perception field (i.e. 1.5$\times$). We propose \emph{fractional dilated convolution}. Without a loss of information, we round up the target scale to an integer dilation factor and stretch the input feature map to a size where the perception field meets the requirement.
Specifically, let $s$ denote the stretch scale and let $\mathrm{interp}_s(\cdot)$ denote a resizing interpolation function (\textit{i.e.}, bilinear interpolation) with scale $s$. We upsample the feature $\bm{h}$ with the stretch scale to $\mathrm{interp}_s(\bm{h})$. The new re-dilation supporting fractional dilation factors is computed as follows:
\begin{equation}
f_{\bm{k}}^d(\bm{h}) = \mathrm{interp}_{1/s}\left(\mathrm{interp}_s(\bm{h}) \circledast \Phi_{\lceil d \rceil} (\bm{k})\right), \;
    s = \lceil d \rceil / d,
\end{equation}
where $\lceil \cdot \rceil$ is the round-up operator. A visualization of re-dilated convolution is shown in Fig.~\ref{fig:re-dilate}. Considering the properties of the diffusion model with $T$ timesteps and $L$ layers, we further generalize the re-dilation factor to become layer-and-timestep-aware, yielding $d = D(t, l)$, where $t \in [0, T-1], l \in [0, L-1]$, and $D$ is a pre-defined dilation schedule function.
Empirically, we find that the re-dilation achieves better synthesis quality when the dilation radius is progressively decreased from deep layers to shallow layers, as well as from noisier steps to less noisy steps, than the fixed dilation factor across all timesteps and layers. 

\begin{figure}[th]
  \centering
  \includegraphics[width=0.9\textwidth]{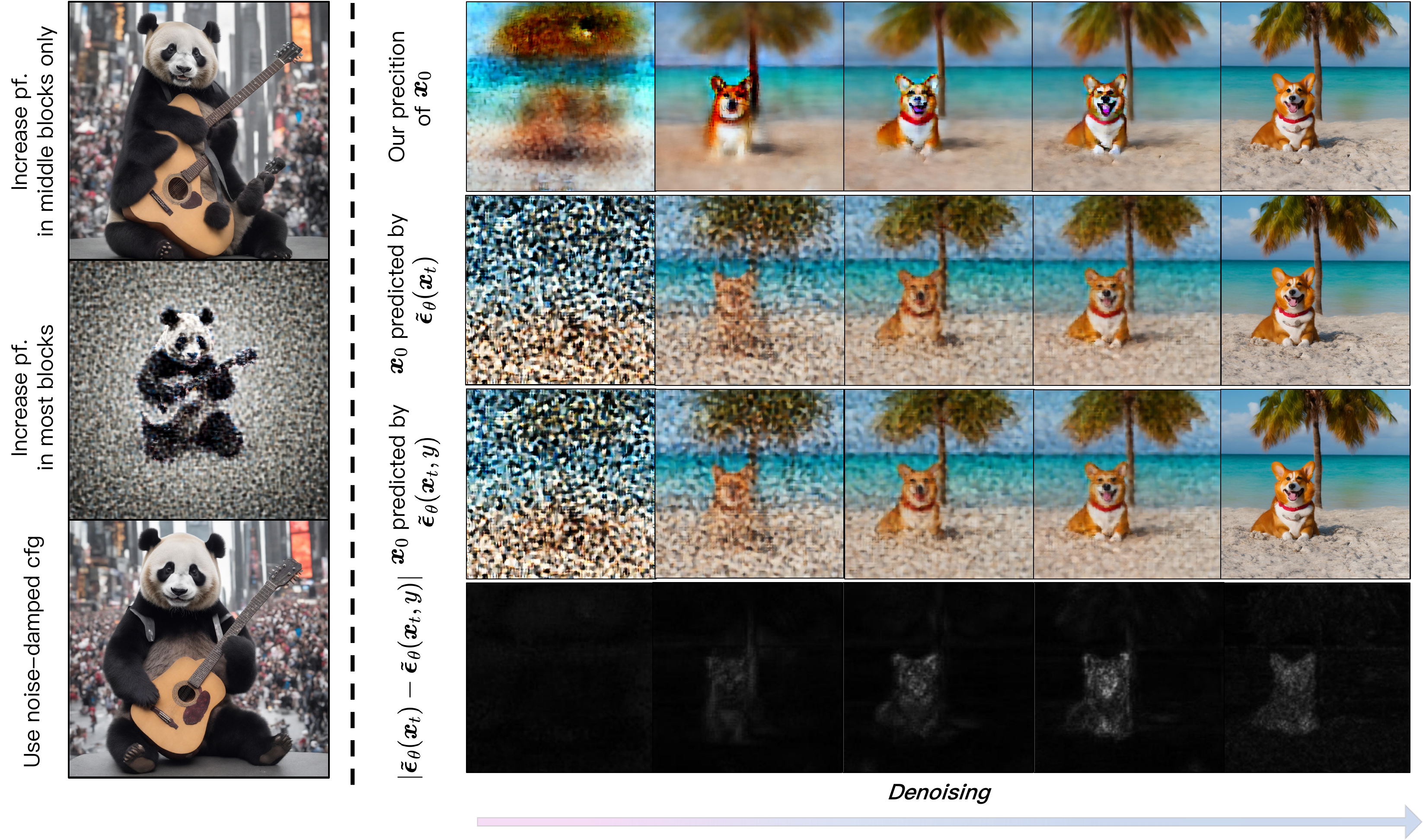}
  \caption{
  \textbf{left}: Samples by increasing perception field in middle blocks and most blocks (middle and outskirt blocks). The middle blocks-only setting fails to produce the correct small object structures. \textbf{right}: The first row shows the predicted original sample using noise-damped classifier-free guidance. The second and third rows show the prediction using $\tilde{\bm{\epsilon}}_\theta(\bx_t, y)$ and $\tilde{\bm{\epsilon}}_\theta(\bx_t)$. $\tilde{\bm{\epsilon}}_\theta(\bx_t, y)$ and $\tilde{\bm{\epsilon}}_\theta(\bx_t)$ fails to remove noise during sampling. However, their predictions exhibit a very similar noise pattern. The fourth row illustrates $\vert \tilde{\bm{\epsilon}}_\theta(\bx_t, y) - \tilde{\bm{\epsilon}}_\theta(\bx_t)\vert$. The erroneous noise prediction vanishes and we can utilize the remaining useful information.
  }
  \label{fig:ndcfg-method}
\end{figure}

\subsection{Convolution Dispersion}
Unfortunately, re-dilated convolution suffers from the periodic downsampling problem (e.g., grinding artifacts) \citep{smoothed-dilation}, \textit{i.e.,} the features will not consider information from a different dilated convolution split. 
This problem arises when adapting a diffusion model to generate much higher resolution. 
To alleviate the problem, we propose to increase the receptive field of a pre-trained convolution layer by dispersing its convolution kernel. Our \emph{convolution dispersion} method is shown in Fig.~\ref{fig:re-dilate}. Given a convolution layer with kernel $\bm{k} \in \mathbb{R}^{r\times r}$ and a target kernel size $r'$ (if the required perception field multiple is $d$ and r is odd, then $r'=d(r - 1) + 1$), our method applies a linear transform $\mathbf{R} \in \mathbb{R}^{r'^2 \times r^2}$ to get a dispersed kernel $\bm{k}' = \mathbf{R} \bm{k}$. We apply \emph{structure-level} and \emph{pixel-level calibration} to enlarge the convolution kernel while keeping the capability of the original convolution layer. 

We use \emph{structure-level calibration} to preserve the performance of a pre-trained convolution layer when the size of the input feature map changes. Consider an arbitrary convolution layer $f_{\bm{k}}(\cdot)$ and the input feature map $\bm{h}$. Structure-level calibration requires the following equation:
\begin{equation}
\label{structure-level-calibration}
\mathrm{interp}_d(f_{\bm{k}}(\bm{h})) = f_{\bm{k}'}(\mathrm{interp}_d(\bm{h})),\; \bm{k}' = \mathbf{R} \bm{k}
\end{equation}
where $f_{\bm{k}'}(\cdot)$ receives the interpolated feature map and keeps its output the same as the interpolated original output $\mathrm{interp}_d(f_{\bm{k}}(\bm{h}))$. Eqn.~\ref{structure-level-calibration} is underdetermined since the enlarged kernel $\bm{k}'$ has more elements than $\bm{k}$. To solve this equation, we introduce \emph{pixel-level calibration} to ensure the enlarged new convolution kernel behaves similarly on the original feature map $\bm{h}$. Mathematically, pixel-level calibration requires $f_{\bm{k}}(\bm{h}) = f_{\bm{k}'}(\bm{h})$. Then, we combine this with Eqn.~\ref{structure-level-calibration} to formulate a linear least square problem:
\begin{equation}
    \mathbf{R} = \min_{\mathbf{R}} \|\mathrm{interp}_d(f_{\bm{k}}(\bm{h})) - f_{\bm{k}'}(\mathrm{interp}_d(\bm{h}))\|_2^2 + \eta \cdot \|f_{\bm{k}}(\bm{h}) - f_{\bm{k}'}(\bm{h})\|_2^2
\end{equation}
where $\eta$ is a weight controlling the focus of dispersed convolution. We derive an enlarged kernel with convolution dispersion by solving the least square problem. Note that $\mathbf{R}$ is not relevant to the exact numerical value of the input feature or the kernel. One can apply it to any convolution kernels to enlarge it from $r\times r$ to $r'\times r'$. Convolution dispersion can be used along with the re-dilation technique to achieve a much larger perceptual field without suffering from the periodic sub-sampling problem. To achieve a fractional perception field scale factor, we replace the dilation operation with convolution dispersion in fractional dilated convolution introduced above.

\subsection{Noise-damped Classifier-free Guidance}

To sample at a much higher resolution (\textit{i.e.}, 4$\times$ in both height and width), we need to increase the perception field in the outer blocks in the denoising U-Net to generate the correct structure in small objects as shown in Fig.~\ref{fig:ndcfg-method}. 
However, we find that the outside block in the U-Net contributes a lot to estimating the noise contained in the input. 
When we try to increase the convolution perceptual field in these blocks, the denoising capability of the model is damaged. 
As a result, it is challenging to generate the correct small structures while maintaining the denoising ability of the original model. 

We propose \emph{noise-damped classifier-free guidance} to address the difficulties. Our method incorporates the two model priors, a model with strong denoising capabilities $\bm{\epsilon}_\theta(\cdot)$, and a model that uses re-dilated or dispersed convolution in most blocks that generates great image content structures $\tilde{\bm{\epsilon}}_\theta(\cdot)$. Then, the sampling process is performed using a linear combination of the estimations with guidance scale $w$:
\begin{equation}
    \label{noise-damped-cfg}
    \bm{\epsilon}_\theta(\bx_t) + w \cdot (\tilde{\bm{\epsilon}}_\theta(\bx_t, y) - \tilde{\bm{\epsilon}}_\theta(\bx_t)),
\end{equation}
where $y$ is the input text prompt. Eqn.~\ref{noise-damped-cfg} includes a base prediction $\bm{\epsilon}_\theta(\bx_t)$ that ensures effective denoising during the sampling process. The guidance term $\tilde{\bm{\epsilon}}_\theta(\bx_t, y) - \tilde{\bm{\epsilon}}_\theta(\bx_t)$ includes two poor noise predictions. 
However, in Fig.~\ref{fig:ndcfg-method}, our experiments demonstrate that the erroneous noise predictions in $\tilde{\bm{\epsilon}}_\theta(\mathbf{x}_t)$ and $\tilde{\bm{\epsilon}}_\theta(\mathbf{x}_t, y)$ are very similar. 
Such erroneous noise prediction vanishes in the results of $\tilde{\bm{\epsilon}}_\theta(\bx_t, y) - \tilde{\bm{\epsilon}}_\theta(\bx_t)$, and the remaining information is useful for generating correct object structures. 




\section{Experiments}




\begin{table}[t]
    \centering
    \resizebox{1.0\linewidth}{!}{
    \begin{tabular}{lccccccccccccc}
    \toprule
        Res & Method & \multicolumn{4}{c}{SD 1.5} & \multicolumn{4}{c}{SD 2.1} & \multicolumn{4}{c}{SD XL 1.0} \\
        \midrule
        & & FID$_{r}$ & KID$_{r}$ & FID$_{b}$ & KID$_{b}$ & FID$_{r}$ & KID$_{r}$ & FID$_{b}$ & KID$_{b}$ & FID$_{r}$ & KID$_{r}$ & FID$_{b}$ & KID$_{b}$\\
        \midrule
        %
        \multirow{2}{*}{4$\times$ 1:1} & Direct-Inf & 38.50 & 0.014 & 29.30 & 0.008 & 29.89 & 0.010 & 24.21 & 0.007 & 67.71 & 0.029 & 45.55 & 0.014 \\
        & Attn-SF & 38.59 & 0.013 & 29.30 & 0.008 & 28.95 & 0.010 & 22.75 & 0.007 & 68.93 & 0.028 & 46.07 & 0.013 \\
        & Ours & \textbf{32.67} & \textbf{0.012} & \textbf{24.93} & \textbf{0.007} & \textbf{20.88} & \textbf{0.008} & \textbf{16.67} & \textbf{0.005} & \textbf{64.75} & \textbf{0.024} & \textbf{28.15} & \textbf{0.009} \\
        \midrule
        \multirow{2}{*}{6.25$\times$ 1:1} & Direct-Inf & 55.47 & 0.020 & 48.54 & 0.015 & 52.58 & 0.018 & 48.13 & 0.014 & 93.91 & 0.041 & 54.90 & 0.020 \\
        & Attn-SF & 55.96 & 0.020 & 49.03 & 0.015 & 50.62 & 0.017 & 45.57 & 0.014 & 93.92 & 0.042 & 54.89 & 0.019 \\
        & Ours & \textbf{52.11} & \textbf{0.019} & \textbf{45.86} & \textbf{0.014} & \textbf{33.36} & \textbf{0.010} & \textbf{30.66} & \textbf{0.008} & \textbf{80.72} & \textbf{0.032} & \textbf{47.15} & \textbf{0.015} \\
        \midrule
        \multirow{2}{*}{8$\times$ 1:2} & Direct-Inf & 74.52 & 0.032 & 68.98 & 0.027 & 69.89 & 0.029 & 55.48 & 0.020 & 122.41 & 0.062 & 82.51 & 0.037 \\
        & Attn-SF & 74.42 & 0.032 & 68.81 & 0.027 & 68.97 & 0.029 & 53.97 & 0.020 & 122.21 & 0.062 & 82.35 & 0.037 \\
        & Ours & \textbf{58.21} & \textbf{0.022} & \textbf{52.76} & \textbf{0.017} & \textbf{58.57} & \textbf{0.021} & \textbf{49.41} & \textbf{0.015} & \textbf{119.58} & \textbf{0.057} & \textbf{50.70} & \textbf{0.019} \\
        \midrule
        \multirow{2}{*}{16$\times$ 1:1} & Direct-Inf & 111.34 & 0.046 & 106.70 & 0.042 & 104.70 & 0.043 & 104.10 & 0.040 & 153.33 & 0.070 & 144.99 & 0.061 \\
        & Attn-SF & 110.10 & 0.046 & 105.42 & 0.042 & 104.34 & 0.043 & 103.61 & 0.041 & 153.68 & 0.070 & 144.84 & 0.061 \\
        & Ours & \textbf{78.22} & \textbf{0.027} & \textbf{65.86} & \textbf{0.023} & \textbf{59.40} & \textbf{0.021} & \textbf{57.26} & \textbf{0.018} & \textbf{131.03} & \textbf{0.063} & \textbf{124.01} & \textbf{0.055} \\
        \bottomrule
    \end{tabular}
    }
    \caption{Quantitative comparisons among training-free methods.}
    \label{tab:quantitative_results}
\end{table}

\paragraph{Experiment setup.} 
%
We conducted evaluation experiments on text-to-image models, Stable Diffusion (SD), including three prevalent versions: SD 1.5~\citep{ldm}, SD 2.1~\citep{sd2-1-base}, and SD XL 1.0~\citep{sdxl} in inferring four unseen higher resolutions.
We experimented with four resolution settings, which are 4 times, 6.25 times, 8 times, and 16 times more pixels than the training. Specifically, for both SD 1.5 and SD 2.1, the training size is 512$^2$, and the inference resolutions are 1024$^2$, 1280$^2$, 2048$\times$1024, 2048$^2$, respectively.
For SD XL, the training resolution is 1024$^2$, and the inference resolutions are 2048$^2$, 2560$^2$, 4096$\times$2048, 4096$^2$.
We also evaluate our approach on a text-to-video model for 2 times higher resolution generation.
Please see our supplementary for the detailed hyperparameter settings.

\paragraph{Testing dataset and evaluation.} We evaluate performance on the dataset of Laion-5B~\citep{laion5b} which contains 5 billion image-caption pairs. When the inference resolution is 1024$^2$, we sample 30k images with randomly sampled text prompts from the dataset. 
Due to massive computation, we sample 10k images when the inference resolution is higher than 1024$^2$. 
In main evaluations, we experiment with a normal higher resolution setting (4$\times$, 1:1), a fractional scaling resolution (6.25, 1:1), a varied aspect ratio (8$\times$, 1:2), and an extreme higher resolution (16$\times$, 1:1). In other experiments, we evaluate the metrics with the aspect ratio of 1:1 unless otherwise specified.
Following the standard evaluation protocol, we measure the Frechet Inception Distance (FID) and Kernel Inception Distance (KID) between generated images and real images to evaluate the generated image quality and diversity, referred to as FID$_r$ and KID$_r$. 
We adopt the implementation of clean-fid~\citep{clean-fid} to avoid discrepancies in the image pre-processing steps.
Since the pre-trained models have the capability of compositing different concepts that do not appear in the training set, we also measure the metrics between the generated samples under the base training resolution and inference resolution, referred to as FID$_\mathrm{b}$ and KID$_\mathrm{b}$.
This evaluates how well our method can preserve the model's original ability when sampling under a new resolution.

\begin{figure}[t]
  \centering
  \includegraphics[width=1.0\textwidth]{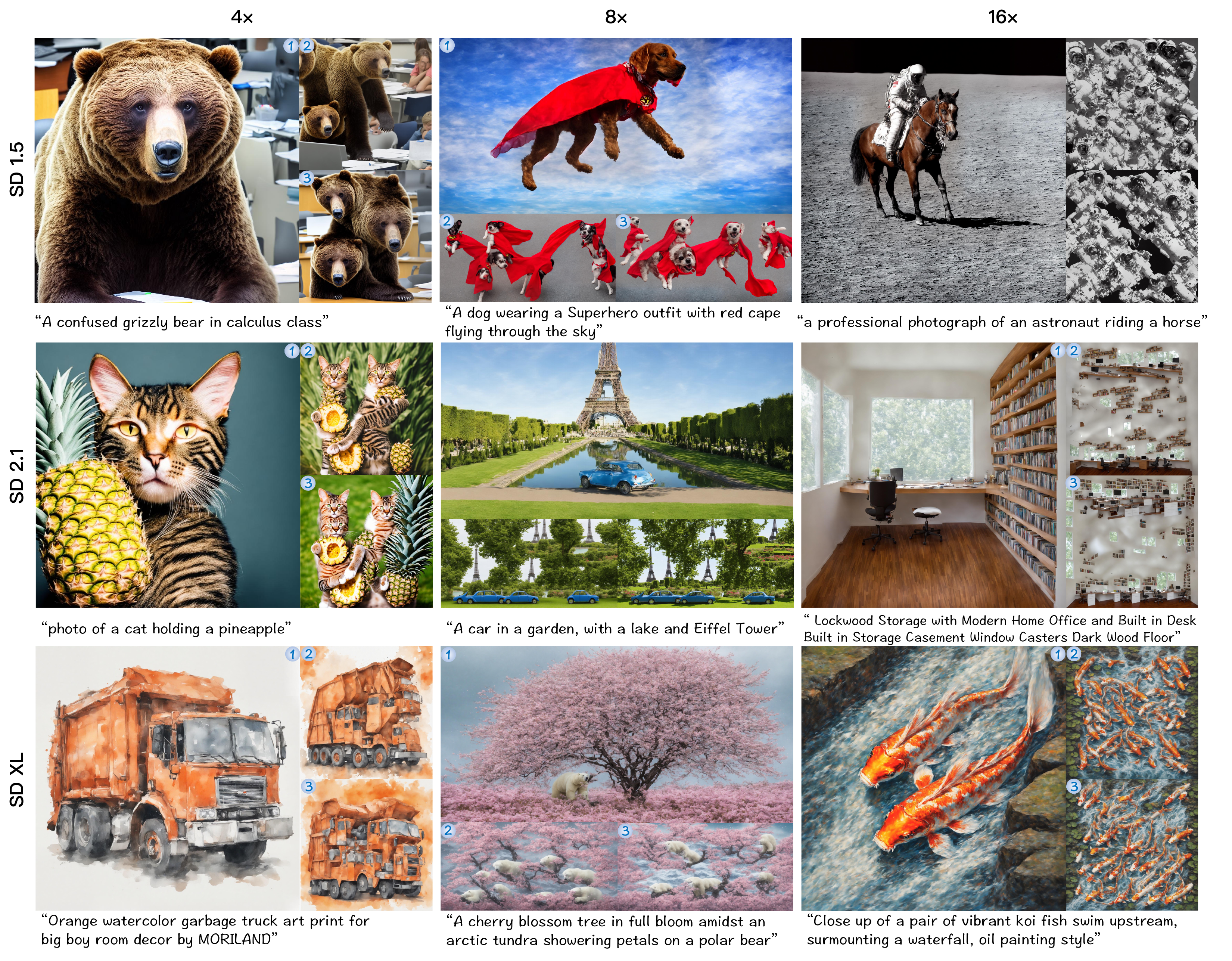}
  \caption{
    Visual comparisons between \ding{172} ours,  \ding{173} directly inferencing SD and  \ding{174} Attn-SF \citep{trainfree-variablesize} in 4$\times$, 8$\times$ and 16$\times$ settings and three Stable Diffusion models. 
  }
  \label{fig:qualitative-compare}
\end{figure}
\begin{figure}[htbp]
    \centering
    \begin{minipage}[t]{0.45\textwidth}
    \centering
    \includegraphics[width=5.9cm]{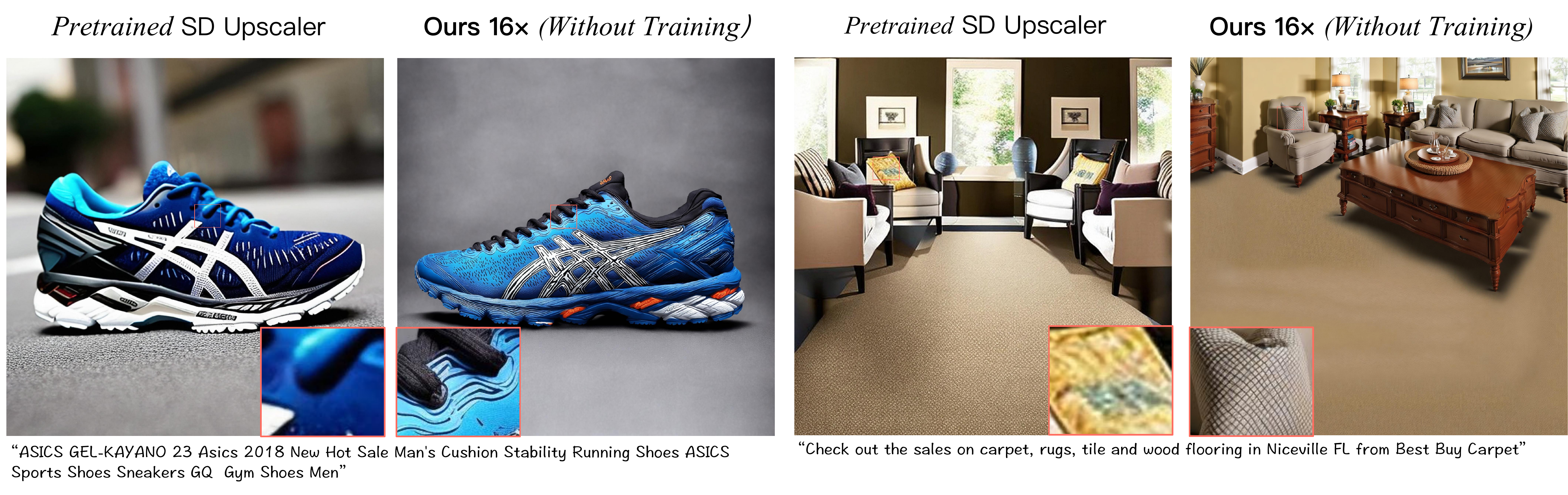}
    \caption{Visual comparisons with SD-SR.}
    \label{fig:compare-sr}
    \end{minipage}
    \begin{minipage}[t]{0.45\textwidth}
    \centering
    \includegraphics[width=5.9cm]{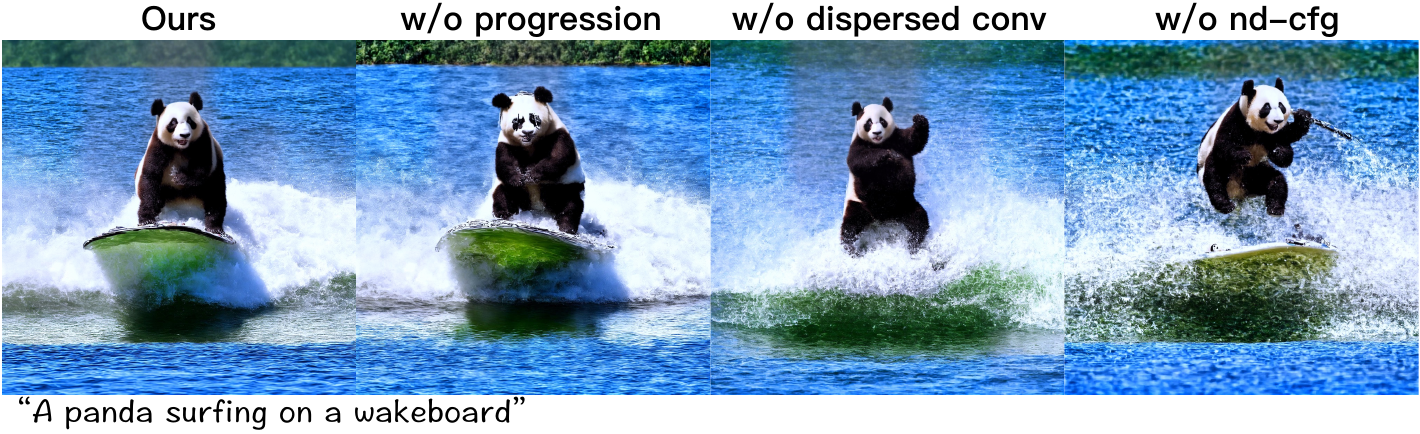}
    \caption{Qualitative ablation results.}
    \label{fig:ablation}
    \end{minipage}
    \vspace{-5mm}
\end{figure}

\subsection{Evaluation}
\paragraph{Comparision with training-free methods.} We compare our method with the vanilla text-to-image diffusion model (Direct-Inf) and a tuning-free method~\citep{trainfree-variablesize} via altering the attention scaling factor (Attn-SF). 
As Multi-Diffusion~\citep{multidiffusion} and SyncDiffusion~\citep{syncdiffusion} cannot alleviate the repetition issue, they are not compared here. 
Our results are shown in Tab.~\ref{tab:quantitative_results}. 
Compared to baselines, we achieve better metric scores in all experiment settings.
It indicates our method preserves the original generation ability of a pre-trained diffusion model much better.
Visual comparisons are shown in Fig.~\ref{fig:qualitative-compare}. 
%

Due to the inappropriate convolution perception field, direct inference and Attn-SF tend to generate small repeated contents, resulting in unnatural image structure. 
On the contrary, our method can generate plausible structures and highly detailed textures in unseen image resolutions. 

\paragraph{Comparison with the diffusion super-resolution model (SR).} 
Although our approach does not require any extra datasets or extensive training efforts. 
To comprehensively evaluate our performance, we compare our approach with a pre-trained Stable Diffusion super-resolution (SD-SR) model: SD 2.1-upscaler-4$\times$~\citep{Upscale} at 4x and 16x higher resolution generation.
Both our approach and the upscaler are combined with the Stable Diffusion 2.1-512$^2$.
Qualitative and quantitative results are shown in \cref{fig:compare-sr} and \cref{tab:compare-sr}.
As seen in \cref{fig:compare-sr}, our method synthesizes better fine-grained details and textures such as the shoes and cushion.
Note that the calculation of FID and KID requires image downsampling to 229 $\times$ 229 which cannot measure the definition of detailed texture.
Hence, we perform a user preference study on the 16 $\times$ setting to ask users to choose an image with a better texture definition between ours and SD Upscaler. The responses of 300 questions are collected. We summarize the results with the percentage of the user's choices, referred to as Texture Definition (TD) (\nth{4} column in the \cref{tab:compare-sr}). 
%
Our FID and KID are slightly worse than the pretrained SD-SR; However, our approach synthesizes high-resolution images with a lower-resolution generative model \textit{in a single stage, and without any extra training}, while the SR model requires a large amount of data and computation to train and tends to exhibit \textit{worse texture details}.
This demonstrates that the pretrained Stable Diffusion model already learns the rich texture priors. With proper utilization, we can leverage this prior to synthesizing high-quality and higher-resolution images.




\begin{figure}[t]
  \centering
  \includegraphics[width=1.0\textwidth]{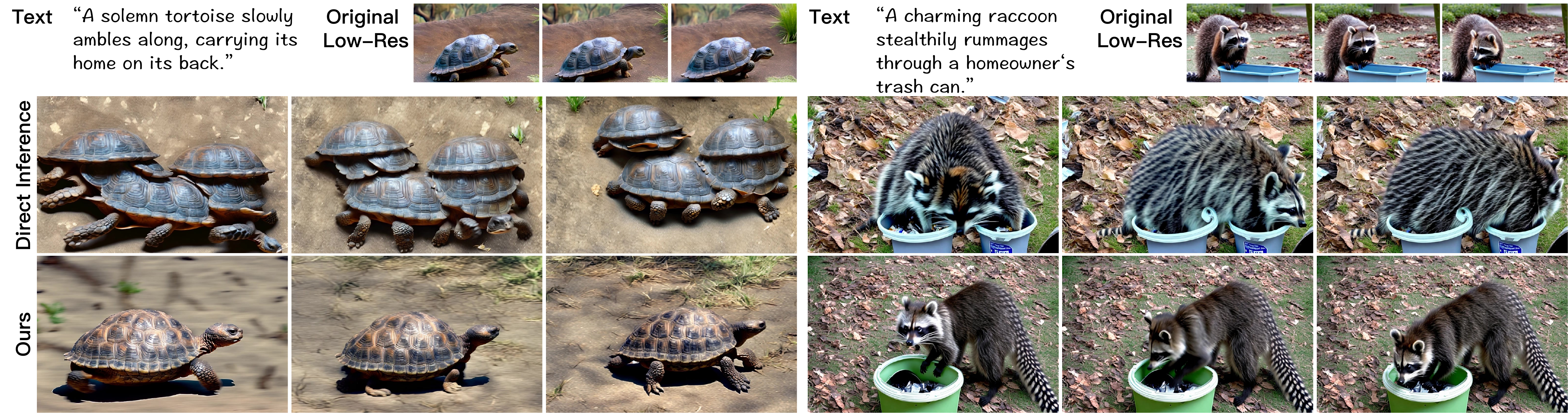}
  \caption{
  Our approach can also be applied to higher-resolution (4x) text-to-video (T2V) generation.
  }
  \label{fig:redilate-video}
\end{figure}

\begin{table}[t]
    \centering
        \begin{minipage}[l]{.47\linewidth}
        \resizebox{1\linewidth}{!}{
        \begin{tabular}{cccc}
        \toprule
            Method & FID$_r$-4$\times$ & KID$_r$-4$\times$ & TD-16$\times$ \\
            \midrule
            SD+SR & 12.59 & 0.005 & 38\% \\
            Ours & 20.88 & 0.008 & \textbf{62\%} \\
            \bottomrule
        \end{tabular}
        }
        \caption{Quantitative comparisons with SR.}
        \label{tab:compare-sr}
        \end{minipage}
        \begin{minipage}[l]{.49\linewidth}
        \resizebox{1\linewidth}{!}{
        \begin{tabular}{ccc}
        \toprule
            Method & FVD$_r$-4$\times$ & KVD$_r$-4$\times$ \\
            \midrule
            Direct Inference & 674.14 & 78.31 \\
            \textbf{Ours} & \textbf{418.80} & \textbf{31.78} \\
            \bottomrule
        \end{tabular}
        }
        \caption{Quantitative results on T2V.}
        \label{tab:dilate-video}
        \end{minipage}
\end{table}

\begin{wraptable}{l}{0.5\linewidth}
    \centering
    \begin{tabular}{ccc}
    \toprule
    Method & FID$_r$-16$\times$ & KID$_r$-16$\times$ \\
    \midrule
    Ours & \textbf{78.22} & \textbf{0.027} \\
    w/o progression & 94.90 & 0.040 \\
    w/o conv dispersion & 106.18 & 0.051 \\
    w/o nd-cfg & 112.15 & 0.055 \\
    \bottomrule
    \end{tabular}
    \caption{Ablation study on SD 1.5.}
    \label{tab:ablation_study}
\end{wraptable}
\subsection{Ablation Study}
We conduct ablation studies of our method on SD 1.5 and 16 $\times$ (1:1) resolution to generate $2048^2$. 
A visual result of removing three key technical components, including timestep-wise progressive re-dilation (progression), convolution dispersion, and noise-damped classifier-free guidance (nd-cfg), can be seen in \cref{fig:ablation}. 
It highlights that the timewise progressive decrease in perception field via re-dilation improves visual quality, especially in image details. 
Noise-damped classifier-free guidance ensures less noise in the final image, and dispersed convolution improves the fidelity of object structures.
Tab.~\ref{tab:ablation_study} shows quantitative results.
Without these components, the FID performance dropped by $16.68$, $27.96$, and $33.93$, respectively. Therefore, every technical component of our method brings considerable improvements.


\subsection{Apply on Video Diffusion Models}
To verify the generalization ability of our method for video generation models, we apply it to a pre-trained text-to-video model, LVDM~\citep{he2022latent}. 
As shown in Fig.~\ref{fig:redilate-video}, our method can generate higher-resolution videos without image definition degeneration.
The quantitative results are in Tab.~\ref{tab:dilate-video}. 
Metrics are computed using the video counterpart Frechet Video Distance (FVD)~\citep{FVD} and Kernel Video Distance (KVD)~\citep{kvd} with $2048$ sampled videos and are evaluated on the Webvid-10M~\citep{webvid}.



\section{Conclusions}
We investigate the possibility of sampling images at a much higher resolution than the training resolution of pre-trained diffusion models. 
Directly sampling a higher resolution image can preserve the image definition but will encounter the severe object repetition issue.
We delve into the architecture of the SD U-Net and explore the receptive field of its components. 
Fortunately, we observe that convolution is critical for sampling higher-resolution images. 
We then propose an elaborate dynamic re-dilation strategy to remove the repetition and also propose the dispersed convolution and noise-damped classifier-free guidance for ultra-high-resolution generation. 
Evaluations are conducted to demonstrate the effectiveness of our methods for different text-to-image and text-to-video models.

\bibliography{ref}

\begin{thebibliography}{34}
\providecommand{\natexlab}[1]{#1}
\providecommand{\url}[1]{\texttt{#1}}
\expandafter\ifx\csname urlstyle\endcsname\relax
  \providecommand{\doi}[1]{doi: #1}\else
  \providecommand{\doi}{doi: \begingroup \urlstyle{rm}\Url}\fi

\bibitem[IF()]{IF}
If.
\newblock URL \url{https://github.com/deep-floyd/IF}.
\newblock Accessed: 9 28, 2023.

\bibitem[Mid()]{Midjourney}
Midjourney.
\newblock URL \url{https://www.midjourney.com/showcase/recent/}.
\newblock Accessed: 9 28, 2023.

\bibitem[Ups()]{Upscale}
Upscaler.
\newblock URL
  \url{https://huggingface.co/stabilityai/stable-diffusion-x4-upscaler}.
\newblock Accessed: 9 29, 2023.

\bibitem[Bain et~al.(2021)Bain, Nagrani, Varol, and Zisserman]{webvid}
Max Bain, Arsha Nagrani, G{\"u}l Varol, and Andrew Zisserman.
\newblock Frozen in time: A joint video and image encoder for end-to-end
  retrieval.
\newblock In \emph{Proceedings of the IEEE/CVF International Conference on
  Computer Vision}, pp.\  1728--1738, 2021.

\bibitem[Bar-Tal et~al.(2023)Bar-Tal, Yariv, Lipman, and Dekel]{multidiffusion}
Omer Bar-Tal, Lior Yariv, Yaron Lipman, and Tali Dekel.
\newblock Multidiffusion: Fusing diffusion paths for controlled image
  generation.
\newblock 2023.

\bibitem[Chang et~al.(2023)Chang, Zhang, Barber, Maschinot, Lezama, Jiang,
  Yang, Murphy, Freeman, Rubinstein, et~al.]{muse}
Huiwen Chang, Han Zhang, Jarred Barber, AJ~Maschinot, Jose Lezama, Lu~Jiang,
  Ming-Hsuan Yang, Kevin Murphy, William~T Freeman, Michael Rubinstein, et~al.
\newblock Muse: Text-to-image generation via masked generative transformers.
\newblock \emph{arXiv preprint arXiv:2301.00704}, 2023.

\bibitem[Chen(2023)]{importance-diffusion}
Ting Chen.
\newblock On the importance of noise scheduling for diffusion models.
\newblock \emph{arXiv preprint arXiv:2301.10972}, 2023.

\bibitem[Diffusion(2022)]{sd2-1-base}
Stable Diffusion.
\newblock Stable diffusion 2-1 base.
\newblock
  \url{https://huggingface.co/stabilityai/stable-diffusion-2-1-base/blob/main/v2-1_512-ema-pruned.ckpt},
  2022.

\bibitem[Ding et~al.(2022)Ding, Zheng, Hong, and Tang]{cogview2}
Ming Ding, Wendi Zheng, Wenyi Hong, and Jie Tang.
\newblock Cogview2: Faster and better text-to-image generation via hierarchical
  transformers.
\newblock \emph{Advances in Neural Information Processing Systems},
  35:\penalty0 16890--16902, 2022.

\bibitem[Gu et~al.(2022)Gu, Chen, Bao, Wen, Zhang, Chen, Yuan, and Guo]{vqdm}
Shuyang Gu, Dong Chen, Jianmin Bao, Fang Wen, Bo~Zhang, Dongdong Chen, Lu~Yuan,
  and Baining Guo.
\newblock Vector quantized diffusion model for text-to-image synthesis.
\newblock In \emph{Proceedings of the IEEE/CVF Conference on Computer Vision
  and Pattern Recognition}, pp.\  10696--10706, 2022.

\bibitem[He et~al.(2022{\natexlab{a}})He, Yang, Zhang, Shan, and
  Chen]{he2022latent}
Yingqing He, Tianyu Yang, Yong Zhang, Ying Shan, and Qifeng Chen.
\newblock Latent video diffusion models for high-fidelity video generation with
  arbitrary lengths.
\newblock \emph{arXiv preprint arXiv:2211.13221}, 2022{\natexlab{a}}.

\bibitem[He et~al.(2022{\natexlab{b}})He, Yang, Zhang, Shan, and Chen]{lvdm}
Yingqing He, Tianyu Yang, Yong Zhang, Ying Shan, and Qifeng Chen.
\newblock Latent video diffusion models for high-fidelity video generation with
  arbitrary lengths.
\newblock \emph{arXiv preprint arXiv:2211.13221}, 2022{\natexlab{b}}.

\bibitem[Ho et~al.(2020)Ho, Jain, and Abbeel]{ddpm}
Jonathan Ho, Ajay Jain, and Pieter Abbeel.
\newblock Denoising diffusion probabilistic models.
\newblock \emph{Advances in Neural Information Processing Systems},
  33:\penalty0 6840--6851, 2020.

\bibitem[Ho et~al.(2022{\natexlab{a}})Ho, Chan, Saharia, Whang, Gao, Gritsenko,
  Kingma, Poole, Norouzi, Fleet, et~al.]{imagen-video}
Jonathan Ho, William Chan, Chitwan Saharia, Jay Whang, Ruiqi Gao, Alexey
  Gritsenko, Diederik~P Kingma, Ben Poole, Mohammad Norouzi, David~J Fleet,
  et~al.
\newblock Imagen video: High definition video generation with diffusion models.
\newblock \emph{arXiv preprint arXiv:2210.02303}, 2022{\natexlab{a}}.

\bibitem[Ho et~al.(2022{\natexlab{b}})Ho, Saharia, Chan, Fleet, Norouzi, and
  Salimans]{cascaded-dm}
Jonathan Ho, Chitwan Saharia, William Chan, David~J Fleet, Mohammad Norouzi,
  and Tim Salimans.
\newblock Cascaded diffusion models for high fidelity image generation.
\newblock \emph{J. Mach. Learn. Res.}, 23:\penalty0 47--1, 2022{\natexlab{b}}.

\bibitem[Hoogeboom et~al.(2023)Hoogeboom, Heek, and Salimans]{simple-diffusion}
Emiel Hoogeboom, Jonathan Heek, and Tim Salimans.
\newblock simple diffusion: End-to-end diffusion for high resolution images.
\newblock \emph{arXiv preprint arXiv:2301.11093}, 2023.

\bibitem[Jin et~al.(2023)Jin, Shen, Li, and Xue]{trainfree-variablesize}
Zhiyu Jin, Xuli Shen, Bin Li, and Xiangyang Xue.
\newblock Training-free diffusion model adaptation for variable-sized
  text-to-image synthesis.
\newblock \emph{arXiv preprint arXiv:2306.08645}, 2023.

\bibitem[Kingma \& Ba(2014)Kingma and Ba]{adam}
Diederik~P Kingma and Jimmy Ba.
\newblock Adam: A method for stochastic optimization.
\newblock \emph{arXiv preprint arXiv:1412.6980}, 2014.

\bibitem[Lee et~al.(2023)Lee, Kim, Kim, and Sung]{syncdiffusion}
Yuseung Lee, Kunho Kim, Hyunjin Kim, and Minhyuk Sung.
\newblock Syncdiffusion: Coherent montage via synchronized joint diffusions.
\newblock \emph{arXiv preprint arXiv:2306.05178}, 2023.

\bibitem[Nichol et~al.(2021)Nichol, Dhariwal, Ramesh, Shyam, Mishkin, McGrew,
  Sutskever, and Chen]{glide}
Alex Nichol, Prafulla Dhariwal, Aditya Ramesh, Pranav Shyam, Pamela Mishkin,
  Bob McGrew, Ilya Sutskever, and Mark Chen.
\newblock Glide: Towards photorealistic image generation and editing with
  text-guided diffusion models.
\newblock \emph{arXiv preprint arXiv:2112.10741}, 2021.

\bibitem[Nichol \& Dhariwal(2021)Nichol and Dhariwal]{im-ddpm}
Alexander~Quinn Nichol and Prafulla Dhariwal.
\newblock Improved denoising diffusion probabilistic models.
\newblock In \emph{International Conference on Machine Learning}, pp.\
  8162--8171. PMLR, 2021.

\bibitem[Parmar et~al.(2022)Parmar, Zhang, and Zhu]{clean-fid}
Gaurav Parmar, Richard Zhang, and Jun-Yan Zhu.
\newblock On aliased resizing and surprising subtleties in gan evaluation.
\newblock In \emph{Proceedings of the IEEE/CVF Conference on Computer Vision
  and Pattern Recognition}, pp.\  11410--11420, 2022.

\bibitem[Podell et~al.(2023)Podell, English, Lacey, Blattmann, Dockhorn,
  M{\"u}ller, Penna, and Rombach]{sdxl}
Dustin Podell, Zion English, Kyle Lacey, Andreas Blattmann, Tim Dockhorn, Jonas
  M{\"u}ller, Joe Penna, and Robin Rombach.
\newblock Sdxl: improving latent diffusion models for high-resolution image
  synthesis.
\newblock \emph{arXiv preprint arXiv:2307.01952}, 2023.

\bibitem[Ramesh et~al.(2022)Ramesh, Dhariwal, Nichol, Chu, and Chen]{dalle-2}
Aditya Ramesh, Prafulla Dhariwal, Alex Nichol, Casey Chu, and Mark Chen.
\newblock Hierarchical text-conditional image generation with clip latents.
\newblock \emph{arXiv preprint arXiv:2204.06125}, 2022.

\bibitem[Rombach et~al.(2022)Rombach, Blattmann, Lorenz, Esser, and Ommer]{ldm}
Robin Rombach, Andreas Blattmann, Dominik Lorenz, Patrick Esser, and Bj{\"o}rn
  Ommer.
\newblock High-resolution image synthesis with latent diffusion models.
\newblock In \emph{Proceedings of the IEEE/CVF Conference on Computer Vision
  and Pattern Recognition}, pp.\  10684--10695, 2022.

\bibitem[Saharia et~al.(2022)Saharia, Chan, Saxena, Li, Whang, Denton,
  Ghasemipour, Ayan, Mahdavi, Lopes, et~al.]{imagen}
Chitwan Saharia, William Chan, Saurabh Saxena, Lala Li, Jay Whang, Emily
  Denton, Seyed Kamyar~Seyed Ghasemipour, Burcu~Karagol Ayan, S~Sara Mahdavi,
  Rapha~Gontijo Lopes, et~al.
\newblock Photorealistic text-to-image diffusion models with deep language
  understanding.
\newblock \emph{arXiv preprint arXiv:2205.11487}, 2022.

\bibitem[Schuhmann et~al.(2022)Schuhmann, Beaumont, Vencu, Gordon, Wightman,
  Cherti, Coombes, Katta, Mullis, Wortsman, Schramowski, Kundurthy, Crowson,
  Schmidt, Kaczmarczyk, and Jitsev]{laion5b}
Christoph Schuhmann, Romain Beaumont, Richard Vencu, Cade Gordon, Ross
  Wightman, Mehdi Cherti, Theo Coombes, Aarush Katta, Clayton Mullis, Mitchell
  Wortsman, Patrick Schramowski, Srivatsa Kundurthy, Katherine Crowson, Ludwig
  Schmidt, Robert Kaczmarczyk, and Jenia Jitsev.
\newblock Laion-5b: An open large-scale dataset for training next generation
  image-text models, 2022.

\bibitem[Teng et~al.(2023)Teng, Zheng, Ding, Hong, Wangni, Yang, and
  Tang]{relay-diffusion}
Jiayan Teng, Wendi Zheng, Ming Ding, Wenyi Hong, Jianqiao Wangni, Zhuoyi Yang,
  and Jie Tang.
\newblock Relay diffusion: Unifying diffusion process across resolutions for
  image synthesis.
\newblock \emph{arXiv preprint arXiv:2309.03350}, 2023.

\bibitem[Unterthiner et~al.(2018)Unterthiner, van Steenkiste, Kurach, Marinier,
  Michalski, and Gelly]{FVD}
Thomas Unterthiner, Sjoerd van Steenkiste, Karol Kurach, Raphael Marinier,
  Marcin Michalski, and Sylvain Gelly.
\newblock Towards accurate generative models of video: A new metric \&
  challenges.
\newblock \emph{arXiv preprint arXiv:1812.01717}, 2018.

\bibitem[Unterthiner et~al.(2019)Unterthiner, van Steenkiste, Kurach, Marinier,
  Michalski, and Gelly]{kvd}
Thomas Unterthiner, Sjoerd van Steenkiste, Karol Kurach, Raphael Marinier,
  Marcin Michalski, and Sylvain Gelly.
\newblock Towards accurate generative models of video: A new metric \&
  challenges.
\newblock \emph{ICLR}, 2019.

\bibitem[Wang \& Ji(2018)Wang and Ji]{smoothed-dilation}
Zhengyang Wang and Shuiwang Ji.
\newblock Smoothed dilated convolutions for improved dense prediction.
\newblock In \emph{Proceedings of the 24th ACM SIGKDD International Conference
  on Knowledge Discovery \& Data Mining}, pp.\  2486--2495, 2018.

\bibitem[Xie et~al.(2023)Xie, Yao, Shi, Liu, Zhou, Liu, Li, and Li]{difffit}
Enze Xie, Lewei Yao, Han Shi, Zhili Liu, Daquan Zhou, Zhaoqiang Liu, Jiawei Li,
  and Zhenguo Li.
\newblock Difffit: Unlocking transferability of large diffusion models via
  simple parameter-efficient fine-tuning.
\newblock \emph{arXiv preprint arXiv:2304.06648}, 2023.

\bibitem[Xing et~al.(2023)Xing, Xia, Liu, Zhang, Zhang, He, Liu, Chen, Cun,
  Wang, et~al.]{make-your-vid}
Jinbo Xing, Menghan Xia, Yuxin Liu, Yuechen Zhang, Yong Zhang, Yingqing He,
  Hanyuan Liu, Haoxin Chen, Xiaodong Cun, Xintao Wang, et~al.
\newblock Make-your-video: Customized video generation using textual and
  structural guidance.
\newblock \emph{arXiv preprint arXiv:2306.00943}, 2023.

\bibitem[Zheng et~al.(2023)Zheng, Guo, Deng, Han, Li, Xu, and
  Xu]{any-size-diffusion}
Qingping Zheng, Yuanfan Guo, Jiankang Deng, Jianhua Han, Ying Li, Songcen Xu,
  and Hang Xu.
\newblock Any-size-diffusion: Toward efficient text-driven synthesis for
  any-size hd images.
\newblock \emph{arXiv preprint arXiv:2308.16582}, 2023.

\end{thebibliography}
\bibliographystyle{iclr2024_conference}

\clearpage
\appendix
\section{Experiment details}
\subsection{Model layers in our method}
The U-Net of Stable Diffusion (SD) v1.5, SD 2.1, and SD XL 1.0 share a similar convolution layer layout. We explain which layer to use re-dilated or dispersed convolutions without a loss of generality. We follow the naming of layers in diffuers\footnote{https://github.com/huggingface/diffusers}. A list of convolution layers contained in a U-Net block is shown in Tab.~\ref{tab:layers-to-dilate}. The attention projection layers and convolution shortcut layers will not use re-dilation or dispersion since the convolution kernel in these layers is 1$\times$1. Note that the first and the last convolution in the U-Net (conv\_in and conv\_out) will not use our method since they do not contribute to generating image contents.
Also, the spatial part of the text-to-video model we used shares the same architecture as the SD. Therefore, layers of the following mentioned are also the same as our video experiment.
\begin{table}[htb]
    \centering
    \resizebox{0.7\linewidth}{!}{
    \begin{tabular}{ccc}
        Layer name & Exist in all blocks & Use our method  \\
        \midrule
        attentions.0.proj\_in & \Checkmark & \XSolidBrush \\
        attentions.0.proj\_out & \Checkmark & \XSolidBrush \\
        attentions.1.proj\_in & \XSolidBrush & \XSolidBrush\\
        attentions.1.proj\_out & \XSolidBrush & \XSolidBrush \\
        attentions.2.proj\_in & \XSolidBrush & \XSolidBrush \\
        attentions.2.proj\_out & \XSolidBrush & \XSolidBrush \\
        resnets.0.conv1 & \Checkmark & \Checkmark \\
        resnets.0.conv2 & \Checkmark & \Checkmark\\
        resnets.0.conv\_shortcut & \XSolidBrush & \XSolidBrush\\
        resnets.1.conv1 & \Checkmark & \Checkmark\\
        resnets.1.conv2 & \Checkmark & \Checkmark\\
        downsamplers.0.conv & \XSolidBrush & \Checkmark \\
        \bottomrule
    \end{tabular}
    }
    \caption{The layers to use our method in a U-Net block. The second column shows the existence condition since some layers cannot be seen in specific U-Net blocks.}
    \label{tab:layers-to-dilate}
\end{table}

\subsection{Hyperparameters}
\begin{figure}[th]
  \centering
  \includegraphics[width=0.75\textwidth]{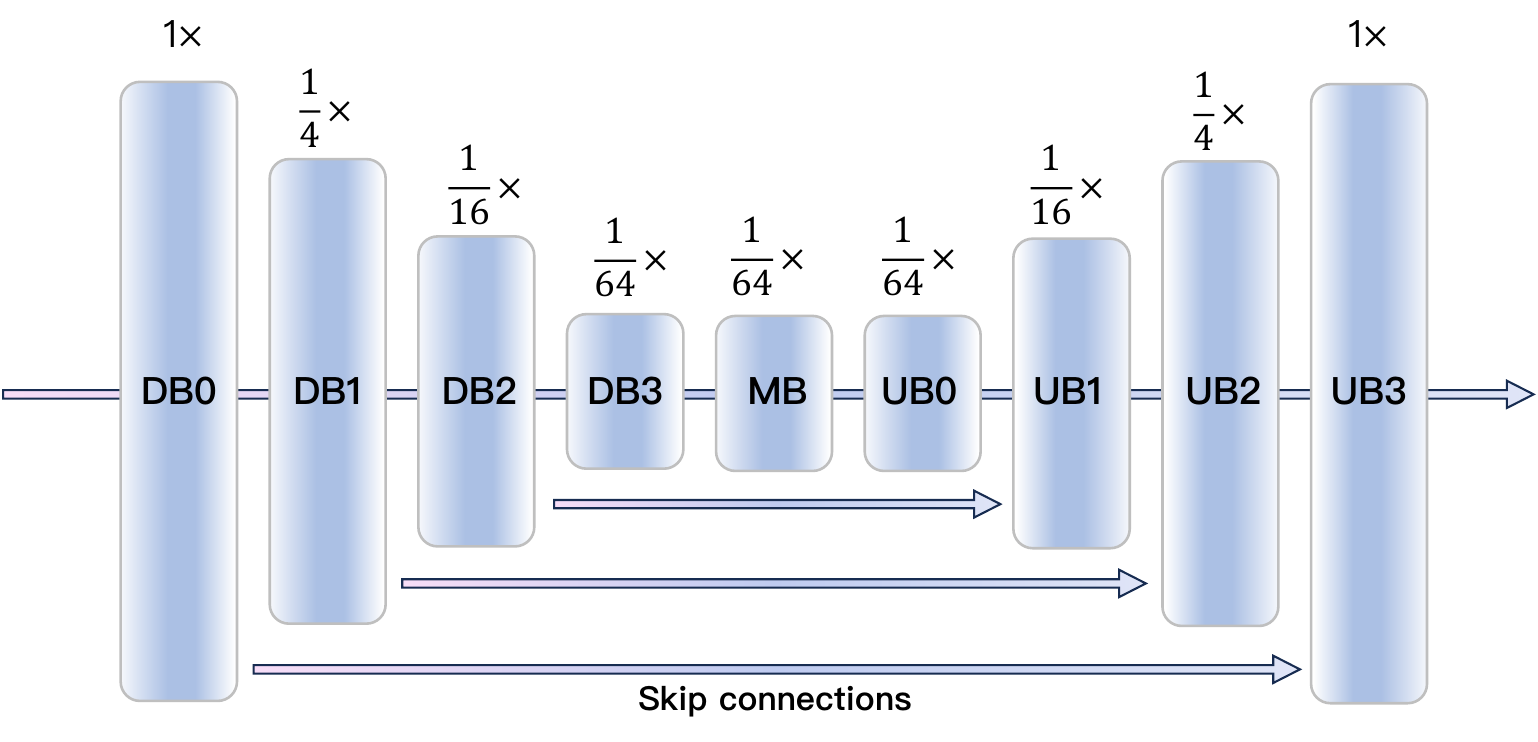}
  \caption{
    Reference block names in the following experiment details. The fractional multiples above blocks are the spatial size of feature maps within the block compared to the network input. i.e, the input latent has 64$^2$ spatial dimension, then the size of feature maps in $\mathrm{MB3}$ is 8$^2$.
  }
  \label{fig:block-names}
\end{figure}
We explain our selection for hyperparameters in this section. All samples are generated using the default classifier-free guidance scale of the corresponding pre-trained model (i.e. SD 1.5 and SD 2.1 use 7.5, SD XL 1.0 uses 5.0). Our SD 2.1 experiments use a similar setting to SD 1.5. We list the hyperparameters for SD 1.5 only for brevity. The evaluation settings for SD 1.5 are shown in Tab.~\ref{tab:sd15-4x}, ~\ref{tab:sd15-625x}, ~\ref{tab:sd15-8x}, ~\ref{tab:sd15-16x}. 
The settings for SD XL 1.0 are shown in Fig.~\ref{tab:sdxl-4x}, ~\ref{tab:sdxl-625x}, ~\ref{tab:sdxl-8x}, ~\ref{tab:sdxl-16x}. A reference for block names and their exact location in the U-Net can be found in Fig.~\ref{fig:block-names}. The tables show detailed settings about which block to use re-dilation conv and dispersed conv. Dilation scale rb. means the dilation scale for re-dilated blocks and dilation scale db. defines the dilation scale for dispersed blocks. If the sampling uses noise-damped classifier-free guidance, we construct a $\bm{\epsilon}_\theta(\cdot)$ with strong denoising capability by turning some outskirt blocks that use re-dilated and dispersed convolution to the original blocks. The chosen ones that become the original blocks are listed in noise-damped blocks. 



\begin{table}[htb]
    \centering
        \begin{minipage}[l]{.49\linewidth}
        \resizebox{1\linewidth}{!}{
        \begin{tabular}{cc}
            Params & Values \\
            \toprule
            latent resolution & 4$\times$128$\times$128 \\
            re-dilated blocks &  $[\mathrm{DB3, MB, UB0}]$\\
            dilation scale rb. & $[2, 2, 2]$\\
            dispersed blocks & $\emptyset$ \\
            progressive & \XSolidBrush \\
            noise-damped cfg. & \XSolidBrush \\
            inference timesteps & 50 \\
            $\tau$ & 30 \\
            \bottomrule
        \end{tabular}
        }
        \caption{1024$^2$ SD 1.5 experiment settings.}
        \label{tab:sd15-4x}
        \end{minipage}
        \begin{minipage}[l]{.49\linewidth}
        \resizebox{1\linewidth}{!}{
        \begin{tabular}{cc}
            Params & Values \\
            \toprule
            latent resolution & 4$\times$160$\times$160 \\
            re-dilated blocks &  $[\mathrm{DB3, MB, UB0}]$\\
            dilation scale rb. & $[2.5, 2.5, 2.5]$\\
            dispersed blocks & $\emptyset$ \\
            progressive & \XSolidBrush \\
            noise-damped cfg. & \XSolidBrush \\
            inference timesteps & 50 \\
            $\tau$ & 30 \\
            \bottomrule
        \end{tabular}
        }
        \caption{1280$^2$ SD 1.5 experiment settings.}
        \label{tab:sd15-625x}
        \end{minipage}
\end{table}



\begin{table}[htb]
    \centering
    \begin{tabular}{cc}
        Params & Values \\
        \toprule
        latent resolution & 4$\times$128$\times$256 \\
        re-dilated blocks &  $[\mathrm{DB0, DB1, DB2, DB3, MB, UB0, UB1, UB2, UB3}]$\\
        dilation scale rb. & $[2, 2, 2, 2, 2, 2, 2, 2, 2]$\\
        dispersed blocks & $\emptyset$ \\
        progressive & \XSolidBrush \\
        noise-damped cfg. & \Checkmark \\
        noise-damped blocks & $[\mathrm{DB0, DB1, DB2, UB1, UB2, UB3}]$ \\
        inference timesteps & 50 \\
        $\tau$ & 30 \\
        \bottomrule
    \end{tabular}
    \caption{2048$\times$1024 SD 1.5 experiment settings.}
    \label{tab:sd15-8x}
\end{table}

\begin{table}[htb]
    \centering
    \begin{tabular}{cc}
        Params & Values \\
        \toprule
        latent resolution & 4$\times$256$\times$256 \\
        re-dilated blocks &  $[\mathrm{DB0, DB1, UB2, UB3}]$\\
        dilation scale rb. & $[2, 4, 4, 2]$\\
        dispersed blocks & $[\mathrm{DB2, DB3, MB, UB0, UB1}]$ \\
        dilation scale db. & $[2, 2, 2, 2, 2]$ \\
        dispersed kernel size & $3 \times 3 \rightarrow 5 \times 5$ \\
        progressive & \Checkmark \\
        noise-damped cfg. & \Checkmark \\
        noise-damped blocks & $[\mathrm{DB0, DB1, UB2, UB3}]$ \\
        inference timesteps & 50 \\
        $\tau$ & 35 \\
        \bottomrule
    \end{tabular}
    \caption{2048$^2$ SD 1.5 experiment settings.}
    \label{tab:sd15-16x}
\end{table}

\begin{table}[htb]
    \centering
        \begin{minipage}[l]{.40\linewidth}
        \resizebox{1\linewidth}{!}{
        \begin{tabular}{cc}
            Params & Values \\
            \toprule
            latent resolution & 4$\times$256$\times$256 \\
            re-dilated blocks &  $[\mathrm{DB3, MB, UB0}]$\\
            dilation scale rb. & $[2, 2, 2]$\\
            dispersed blocks & $\emptyset$ \\
            progressive & \XSolidBrush \\
            noise-damped cfg. & \XSolidBrush \\
            inference timesteps & 50 \\
            $\tau$ & 30 \\
            \bottomrule
        \end{tabular}
        }
        \caption{2048$^2$ SD XL 1.0 settings.}
        \label{tab:sdxl-4x}
        \end{minipage}
        \begin{minipage}[l]{.55\linewidth}
        \resizebox{1.1\linewidth}{!}{
        \begin{tabular}{cc}
            Params & Values \\
            \toprule
            latent resolution & 4$\times$320$\times$320 \\
            re-dilated blocks &  $[\mathrm{DB1, DB2, DB3, MB, UB0, UB1, UB2}]$\\
            dilation scale rb. & $[2, 2, 2.5, 2.5, 2.5, 2, 2]$\\
            dispersed blocks & $\emptyset$ \\
            progressive & \XSolidBrush \\
            noise-damped cfg. & \Checkmark \\
            noise-damped blocks & $[\mathrm{DB1, DB2, UB1, UB2}]$ \\
            inference timesteps & 50 \\
            $\tau$ & 30 \\
            \bottomrule
        \end{tabular}
        }
        \caption{2560$^2$ SD XL 1.0 experiment settings.}
        \label{tab:sdxl-625x}
        \end{minipage}
\end{table}



\begin{table}[htb]
    \centering
        \begin{minipage}[l]{.6\linewidth}
        \resizebox{1\linewidth}{!}{
        \begin{tabular}{cc}
            Params & Values \\
            \toprule
            latent resolution & 4$\times$256$\times$512 \\
            re-dilated blocks &  $[\mathrm{DB1, DB2, DB3, MB, UB0, UB1, UB2}]$\\
            dilation scale rb. & $[2, 2, 2, 2, 2, 2, 2]$\\
            dispersed blocks & $\emptyset$ \\
            progressive & \XSolidBrush \\
            noise-damped cfg. & \Checkmark \\
            noise-damped blocks & $[\mathrm{DB1, DB2, UB1, UB2}]$ \\
            inference timesteps & 50 \\
            $\tau$ & 30 \\
            \bottomrule
        \end{tabular}
        }
        \caption{4096$\times$2048 SD XL 1.0 experiment settings.}
        \label{tab:sdxl-8x}
        \end{minipage}
        \begin{minipage}[l]{.35\linewidth}
        \resizebox{1.0\linewidth}{!}{
        \begin{tabular}{cc}
            Params & Values \\
            \toprule
            latent resolution & 4$\times$512$\times$512 \\
            re-dilated blocks &  $[\mathrm{DB2, UB1}]$\\
            dilation scale rb. & $[2, 2]$\\
            dispersed blocks & $[\mathrm{DB3, MB, UB0}]$ \\
            dilation scale db. & $[2, 2]$ \\
            dispersed kernel size & $3 \times 3 \rightarrow 5 \times 5$ \\
            progressive & \Checkmark \\
            noise-damped cfg. & \Checkmark \\
            noise-damped blocks & $[\mathrm{DB2, UB1}]$ \\
            inference timesteps & 50 \\
            $\tau$ & 35 \\
            \bottomrule
        \end{tabular}
        }
        \caption{4096$^2$ SD XL settings.}
        \label{tab:sdxl-16x}
        \end{minipage}
\end{table}


\subsection{synchronize statistics between tiles in GroupNorm}
\begin{wrapfigure}{i}[0cm]{0.43\linewidth}
  \centering
  \includegraphics[width=0.43\textwidth]{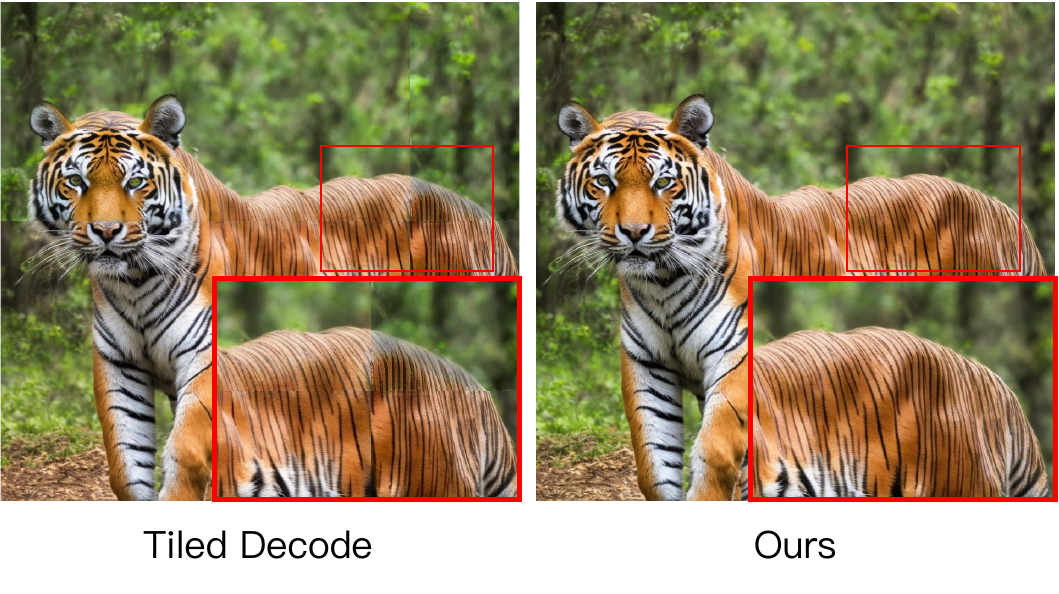}
  \caption{Direct tiled decode causes abrupt changes in tile borders and different color tones in tiles. We synchronize the statistics in VAE GroupNorm between tiles to address this problem.}
  \label{fig:sync-gn}
\end{wrapfigure}

When the generated image size is large (i.e., $>2048\times2048$), the VAE of SD requires enormous VRAM for decoding and is usually not applicable on a personal GPU. A simple solution is decoding in tiles. However, tiled decoding usually causes abrupt changes between different tiles as shown in Fig.~\ref{fig:sync-gn}. To solve this, one can make overlapped regions between tiles and interpolate on the overlapped regions. However, another problem of tiled decoding is the inconsistent color tone between tiles. We figure out this is caused by the independent computation of GroupNorm (GN) layers in VAE between tiles. We propose to synchronize the feature statistics in GN in different tiles. Specifically, we compute the mean and std using all tiles instead of using only current ones. As shown in Fig.~\ref{fig:sync-gn}, it eliminates the color tone difference efficiently.

\section{Re-dilated attention}
\begin{wrapfigure}{i}[0cm]{0.3\linewidth}
  \centering
  \includegraphics[width=0.3\textwidth]{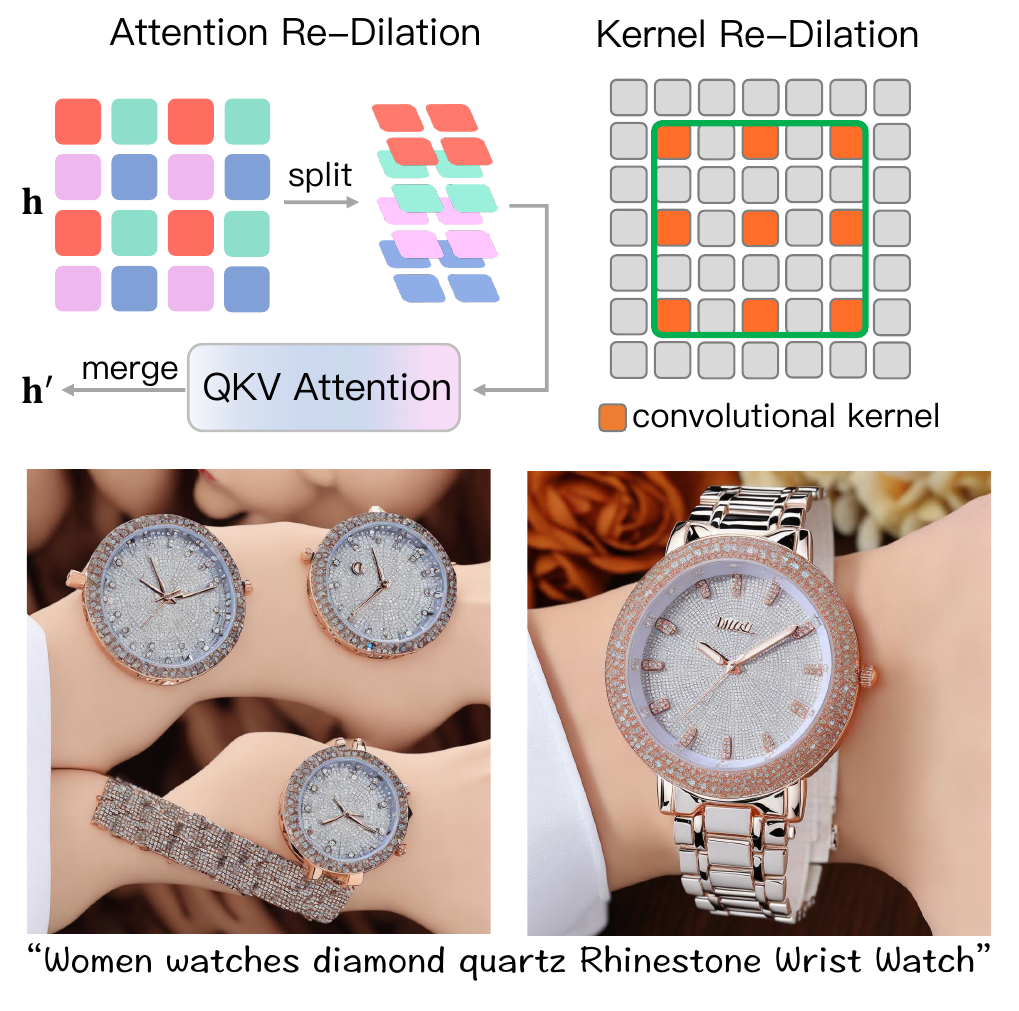}
  \caption{Illustration and results of two re-dilations.
  }
  \label{fig:redilated-attn}
\end{wrapfigure}
Here, we introduce the experimented re-dilated attention.
We aim to keep the original receptive field of attention, e.g., the attention token quantity.
Thus, before calculating the attentional features, we first split the input feature map into four slices (the resolution is 4x higher than the training), and for each slice, we flat them into token sequences and feed them into the QKV attention. 
After the attention calculation, we merge them back to form the original feature arrangement.
This operation strictly controls the token length of attention to be the same as training. However, this cannot solve the structure issue of the generated image, as shown in the \nth{2} row of \cref{fig:redilated-attn}.
However, when applying the redilation on the convolutional kernel, the structure is totally correct.
This demonstrates that the key cause of structure repetition lies in convolutional kernels.

\clearpage
\section{Other visualizations}
\begin{figure}[th]
  \centering
  \includegraphics[width=1.0\textwidth]{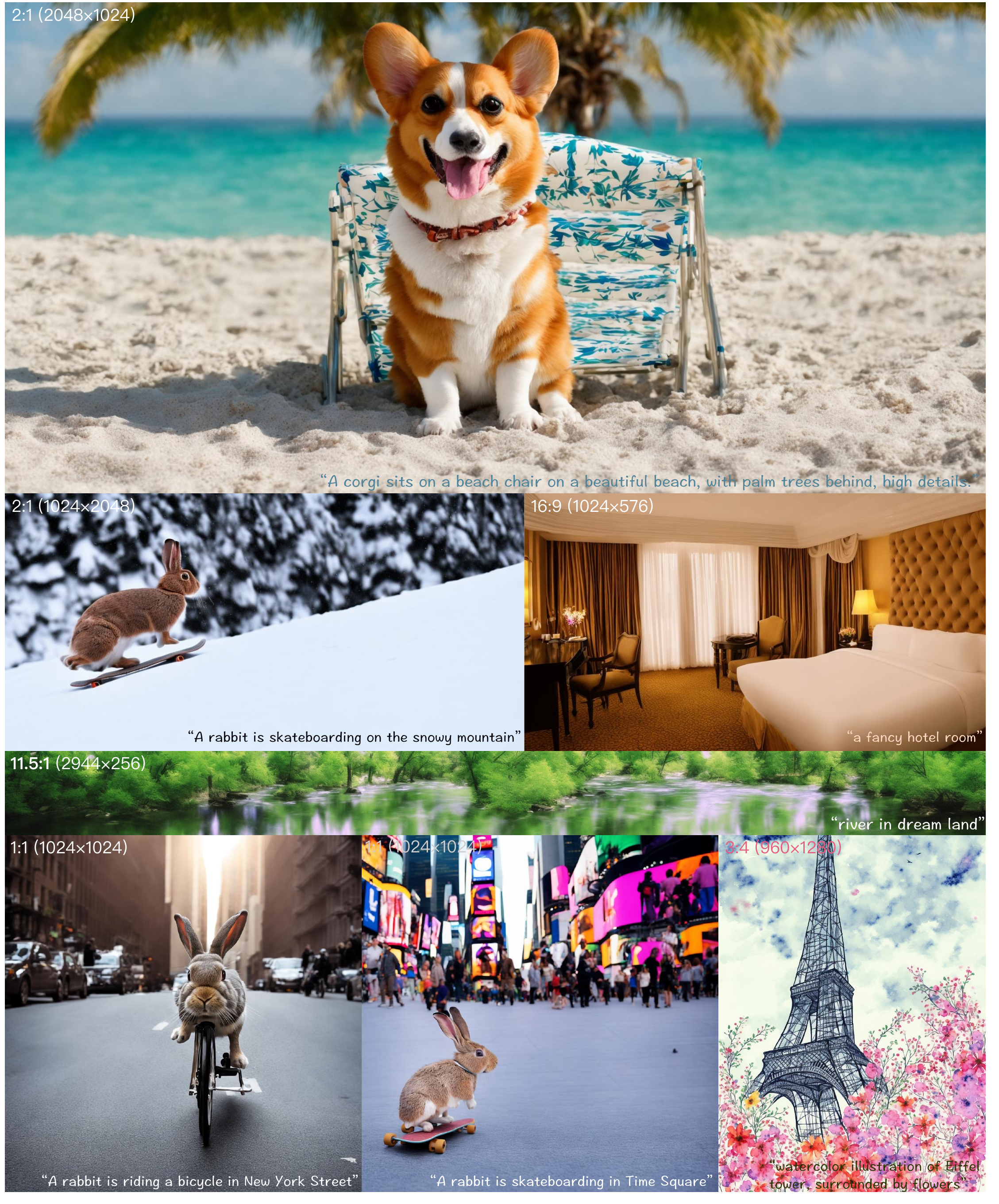}
  \caption{
  More generated results with our method and SD 2.1 with arbitrary aspect ratios and sizes.
  }
  \label{fig:more-res-sd22.1}
\end{figure}
\begin{figure}[th]
  \centering
  \includegraphics[width=1.0\textwidth]{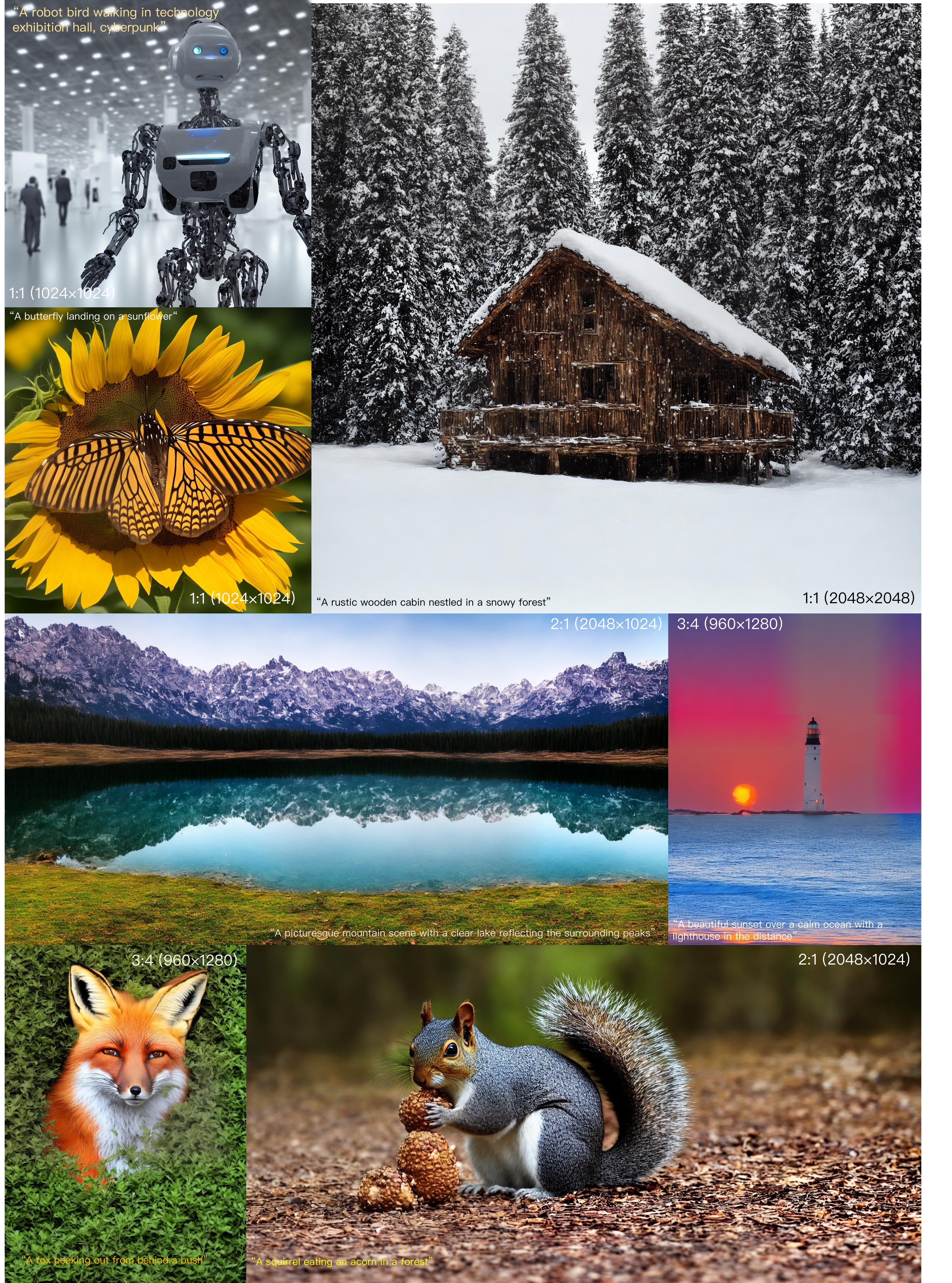}
  \caption{
  More generated results with our method and SD 1.5 with arbitrary aspect ratios and sizes.
  }
  \label{fig:more-res-sd22.1}
\end{figure}

\begin{figure}[th]
  \centering
  \includegraphics[width=1.0\textwidth]{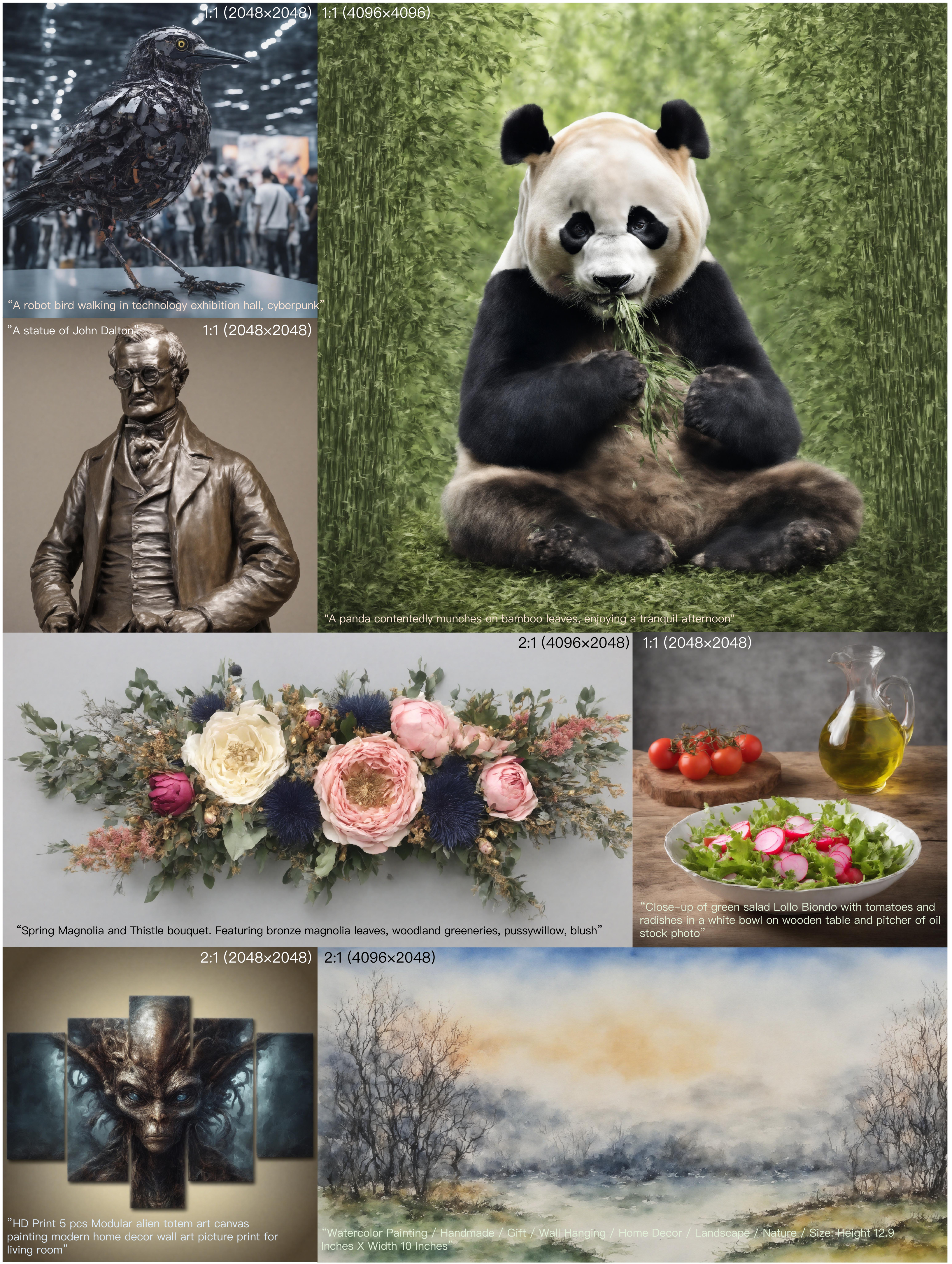}
  \caption{
  More generated results with our method and SD XL 1.0 with arbitrary aspect ratios and sizes.
  }
  \label{fig:more-res-sdxl}
\end{figure}

\end{document}